\pdfoutput=1

\documentclass[11pt]{article}

\usepackage{EMNLP2022}

\usepackage{times}
\usepackage{latexsym}

\usepackage[T1]{fontenc}

\usepackage[utf8]{inputenc}

\usepackage{microtype}

\usepackage{inconsolata}

\usepackage{floatrow}
\newfloatcommand{capbtabbox}{table}[][\FBwidth]
\usepackage{times}
\usepackage{latexsym}

\usepackage{appendix}
\usepackage{textcomp}
\usepackage{listings}
\usepackage{amsmath}

\usepackage{enumitem}
\usepackage{amsmath} 
\usepackage{pifont}
\DeclareMathSymbol{\qm}{\mathalpha}{operators}{"3F}
\DeclareMathAlphabet{\mathbbold}{U}{bbold}{m}{n}
\usepackage{graphicx}
\usepackage{booktabs}
\usepackage{multirow}
\usepackage{makecell}
\usepackage{tablefootnote}
\usepackage{subcaption}
\usepackage{pifont}

\usepackage{nccmath, mathtools}
\usepackage[T1]{fontenc}
\usepackage[scaled=0.7]{beramono} 
\newcommand*\samethanks[1][\value{footnote}]{\footnotemark[#1]}

\title{Dialogue Meaning Representation for Task-Oriented Dialogue Systems}

\author{Xiangkun Hu\textsuperscript{1,}\thanks{\ \  Equal contribution.} , Junqi Dai\textsuperscript{2,}\samethanks\ \  \textsuperscript{,}\thanks{{} {} Work done during internship at Amazon Shanghai AI Lab.} , Hang Yan\textsuperscript{2}, Yi Zhang\textsuperscript{1}, \\ \textbf{Qipeng Guo\textsuperscript{1} }, \textbf{Xipeng Qiu\textsuperscript{2}}, \textbf{Zheng Zhang\textsuperscript{1}}\\
  \textsuperscript{1}Amazon AWS AI \\
  \textsuperscript{2}School of Computer Science, Fudan University \\
  \texttt{\{xiangkhu, yizhngn, gqipeng, zhaz\}@amazon.com} \\
  \texttt{jqdai22@m.fudan.edu.cn} \\
  \texttt{\{hyan19, xpqiu\}@fudan.edu.cn}\\}

\begin{document}
\maketitle
\begin{abstract}
Dialogue meaning representation formulates natural language utterance semantics in their conversational context in an explicit and machine-readable form. Previous work typically follows the intent-slot framework, which is easy for annotation yet limited in scalability for complex linguistic expressions. A line of works alleviates the representation issue by introducing hierarchical structures but challenging to express complex compositional semantics, such as negation and coreference. We propose  Dialogue Meaning Representation (DMR), a pliable and easily extendable representation for task-oriented dialogue. Our representation contains a set of nodes and edges to represent rich compositional semantics. Moreover, we propose  an inheritance hierarchy mechanism focusing on   domain extensibility. Additionally, we annotated DMR-FastFood, a multi-turn dialogue dataset with more than 70k utterances, with DMR. We propose two evaluation tasks to evaluate different dialogue models and a novel coreference resolution model GNNCoref for the graph-based coreference resolution task. Experiments show that DMR can be parsed well with pre-trained Seq2Seq models, and GNNCoref outperforms the baseline models by a large margin.\footnote{The dataset and code are available at \url{https://github.com/amazon-research/dialogue-meaning-representation}.}

\end{abstract}

\section{Introduction}

A task-oriented dialogue (TOD) system aims to serve users to accomplish tasks in specific domains with interactive conversations. The modeling of converting natural language semantics in their conversational context into a machine-readable structured representation, also known as the dialogue meaning representation, is at the core of TOD system research. A meaning representation framework sets the stage for natural language understanding (NLU) and allows the system to communicate with other downstream components such as databases or web service APIs.

\begin{figure}[t!]
    \centering
    \includegraphics[width=\textwidth]{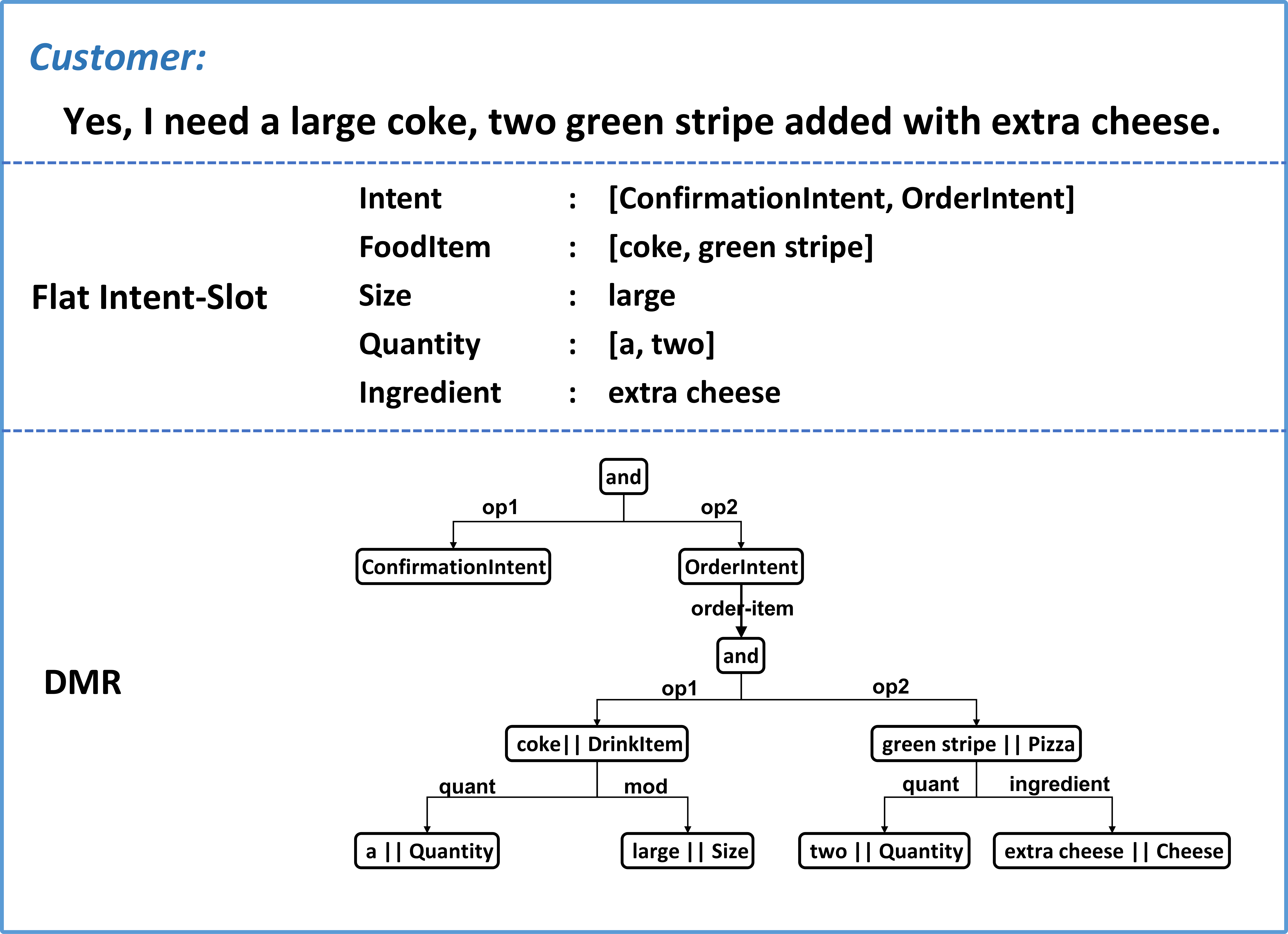}
    \caption{An example of a customer's utterance for food ordering.   The  flat intent-slot schema  can not align the food items (``coke'' and ``green stripe'') with the modifiers (``large'', ``a'', ``two'' and ``extra cheese'') in conjunction constructions. These multiple conjunctions and modifiers require a good meaning representation to reveal the relations between attributes (e.g., size, quantity) and their corresponding entities.  Here, we propose DMR with an example shown in the lower part of the figure, a meaning representation for TOD, which can resolve such compositional semantics.
}
    \label{fig:case}
\end{figure}

Derived from the theoretical framework of \citet{fillmore:1968}, and wildly adopted in dialogue system designs as early as \citet{bobrow:1977}, the classic flat intent-slot schema represents an utterance into one specific intent with several associated slots. Such schemas are convenient in annotation but limited in the expression for compositional semantics,    such as conjunction, modification, negation, and coreference across dialogue turns. These complex patterns are not at all uncommon in real-world use cases. Take the fast-food domain data from  MultiDoGO dataset~\citep{peskov-etal-2019-multi} as an example, where an agent is required to extract information about ordering food (e.g., food name, quantity, size, and ingredients) from the conversation with the customer, about 12.3\% of the utterances contain multi-intent and 11.2\% are involved with coreference semantics.   However, the flat intent-slot schema leaves these semantics uncovered. Moreover, 8.7\% of the ordering utterances contain multiple objects and modifiers. For example in \figurename~\ref{fig:case}, with the flat intent-slot schema, the extracted food items ``coke'' and ``green stripe'' (a pizza name) cannot be aligned with the size ``large'', the quantities ``a'' and ``two'', and the ingredient ``extra cheese''.

The vanilla flat intent-slot schema needs to be extended to support compositional semantics. \citet{DBLP:conf/emnlp/GuptaSMKL18} proposes TOP in the form of a hierarchical parsing tree to allow representation for nested intents. Its successor, SB-TOP \cite{DBLP:conf/emnlp/AghajanyanMSDHL20}, further simplifies the structure and  supports  coreference. \citet{DBLP:conf/emnlp/ChengAABDFKKLPW20} introduces TreeDST, which is also a tree-structured dialogue state representation to allow high compositionality and integrate domain knowledge. These studies imply the trend pointing to a   balance in a better expression for complex compositional semantics  and a lighter structure for extensibility.

This paper proposes Dialogue Meaning Representation (DMR) that significantly extends the intent-slot framework. DMR is a rooted, directed acyclic graph (DAG)     composed of nodes of \emph{Intent}, \emph{Entity} and pre-defined \emph{Operator} and \emph{Keyword}, as well as edges between them. Entity is an extension of slot which wraps the slot value with the specific slot type defined in external knowledge. Such design allows arbitrary complex compositionality between slots and keeps the potential for type constraint. Operator and Keyword are components to represent linguistic knowledge (i.e. general semantics) such as conjunction, negation, quantification, coreference, etc. The details of DMR can be found in Section \ref{dmr_desc}, and the example comparing DMR with flat intent-slot representation is shown in \figurename~\ref{fig:case}. As described later in Section  \ref{dmr_desc}, many of the key designs are inspired by AMR\cite{DBLP:conf/acllaw/BanarescuBCGGHK13} but specialized for TOD. Thus, DMR can be considered a dialect of AMR. From this perspective, DMR is powerful enough and easily extendable for TOD applications. 

Moreover, DMR is capable of adapting to different domains. Unlike previous works, DMR utilizes a  domain-agnostic ontology to define the structural constraints and representations of general semantics. It allows chatbot developers to derive domain-specific ontology from this for their applications through the \emph{Inheritance Hierarchy} mechanism. This design improves both generalization and normalization of DMR. 


To validate our idea, we propose a dataset, DMR-FastFood, with 7194 dialogues and 70328 annotated utterances. This dataset is extensively annotated with more linguistic semantics, including 16087 conjunctions and 557 negations, significantly more than other related datasets. We developed and evaluated a few baseline models to pinpoint where the challenges lie. We further propose GNNCoref for the coreference resolution task on DMR. In general, DMR parsing is not difficult, especially with a pre-trained model, though graphs with more complex (and often deeper) structures are naturally more challenging. Moreover, experiments show that GNNCoref performs better compared to baseline models. 

\section{Dialogue Meaning Representation}
\label{dmr_desc}
This section describes the structure of the DMR graph, the domain-agnostic ontology, and the representation of general semantics. 


\subsection{DMR Ontology}

\begin{figure}[ht!]
    \centering
    \includegraphics[width=\columnwidth]{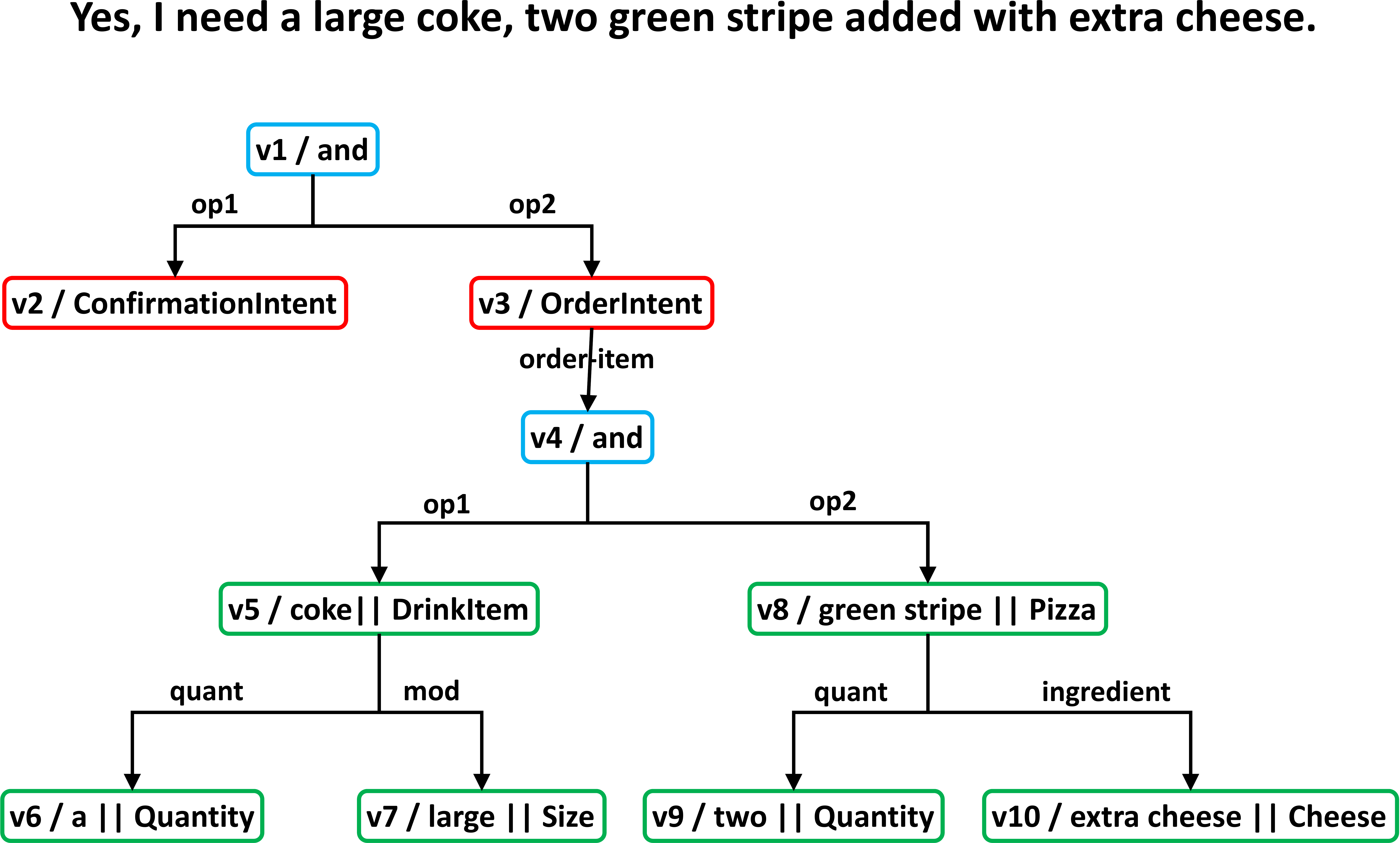}
    \caption{%
        An example of DMR graph from fast food domain. Nodes in different types are in different box colors, namely,   {\color[RGB]{255,28,3}{\texttt{Intent}}},  {\color[RGB]{91,155,213}{\texttt{Operator}}} and {\color[RGB]{112,173,71}{\texttt{Entity}}}.
    }
    \label{fig:nodes}
\end{figure}

DMR ontology defines the nodes, the edges, and the rules for constructing DMR graphs. It also describes the inheritance hierarchy mechanism. DMR is a rooted directed acyclic graph with node and edge labels. Figure~\ref{fig:nodes} shows an example of a DMR graph from the fast-food domain.

\subsubsection{Nodes}
Different from the general purpose of the predicates and concepts defined in AMR, DMR utilizes the specialized designed nodes for TOD.
There are four types of nodes in DMR:

\begin{itemize}
    \item \texttt{Intent}, denotes the intention of the speaker, such as \texttt{OrderIntent} in Figure~\ref{fig:nodes};
    
    \item \texttt{Entity}, denotes the objects mentioned in the utterance. Generally, it is formed with ``\texttt{<lexical\_value> $\Vert$ <canonical\_value> $\Vert$ Entity}'', where \texttt{<lexical\_value>} specifies the surface form value  from the utterance, and \texttt{<canonical\_value>} is the predefined value for the entity in ontology.  \texttt{<lexical\_value>} and \texttt{<canonical\_value>} are optional;\footnote{In our DMR-FastFood dataset, there are no pre-defined canonical values, we instead use the form \texttt{``<lexical\_value> $\Vert$ Entity''} for simplicity.}
    
    \item \texttt{Operator}, supports compositional constructions, such as \texttt{and} for conjunction, and \texttt{reference} for cross-turn coreference. Details are described in the next section;
    
    \item \texttt{Keyword}, specifies keywords for some special semantics, such as ``\texttt{-}'' for negation. 
\end{itemize}

Each node (except keywords) are assigned with an identifier as AMR does, such as ``v1'' for node \texttt{OrderIntent} in Figure~\ref{fig:nodes}. And the root node of a DMR graph is restricted to be an \texttt{Intent} node or a conjunction operator \texttt{and} for packing multiple intentions.





\subsubsection{Edges}



The nodes in DMR graph are linked with directed edges.  For every edge, the  node types it can reach are pre-defined. Moreover, all types of nodes have pre-defined arguments in the ontology,  which constrains the argument type, namely the edge here, and node types the edge can reach. In a specific DMR graph,  some arguments defined in  ontology  may not appear. For example, in fast food domain,  intent \texttt{OrderIntent} has one argument \texttt{order-item} (see Figure~\ref{fig:nodes}). 
Entity type \texttt{DrinkItem} has pre-defined arguments \texttt{quant}, \texttt{mod} and \texttt{ingredient}, but in the example of Figure~\ref{fig:nodes}, no ingredients of the coke are mentioned in the utterance, thus, the edge \texttt{ingredient} is not shown in the graph.

\subsubsection{Inheritance Hierarchy}
DMR is featured with inheritance hierarchy. With this mechanism, chatbot developers can derive domain-specific ontology easily and organize it hierarchically. For example, in the fast-food domain, we can derive the ontology like:

\begin{lstlisting}[mathescape=true]
Intent $\leftarrow$ OrderIntent | PaymentIntent | ThankYouIntent
Entity $\leftarrow$ FoodItem | DrinkItem
FoodItem $\leftarrow$ Pizza | Burger | Sandwich
\end{lstlisting} 
It defines three intents \texttt{OrderIntent}, \texttt{PaymentIntent} and \texttt{ThankYouIntent}; two base entity types \texttt{FoodItem} and \texttt{DrinkItem}; and three \texttt{FoodItem} types \texttt{Pizza}, \texttt{Burger} and \texttt{Sandwich} that inherits from \texttt{FoodItem}. The derived intents and entity types inherit their parents' arguments by default. In the above example, \texttt{FoodItem} and \texttt{DrinkItem} inherit arguments of \texttt{Entity} such as \texttt{mod} and \texttt{quant}; \texttt{Pizza}, \texttt{Burger} and \texttt{Sandwich} inherit arguments of \texttt{FoodItem} such as \texttt{ingredient}, and so on. We describe more details on this in Appendix~\ref{appendix:inheritance}.

With inheritance hierarchy, the domain-specific and domain-agnostic knowledge are well separated: the general semantics that is common in all domains, such as  quantification, and negation, are defined in the domain-agnostic ontology, while the domain-specific part inherits these representations and can focus on the application. Further, it reduces the burden of constructing ontology, as the intent and entity types inherit their parents' arguments by default, and the ontology is organized hierarchically.



\subsection{Compositional Semantics}
\label{2.2}

Here we describe the general compositional semantics defined in the domain-agnostic ontology. It is worth noting that we do not cover all the general semantics in TOD, though this set can be extended in the future. The examples used are taken from the fast-food domain; we just show a sub-graph of the DMRs and omit the variables for simplicity.

\begin{figure}[h!]
    \centering
    \strut
    \includegraphics[width=\columnwidth]{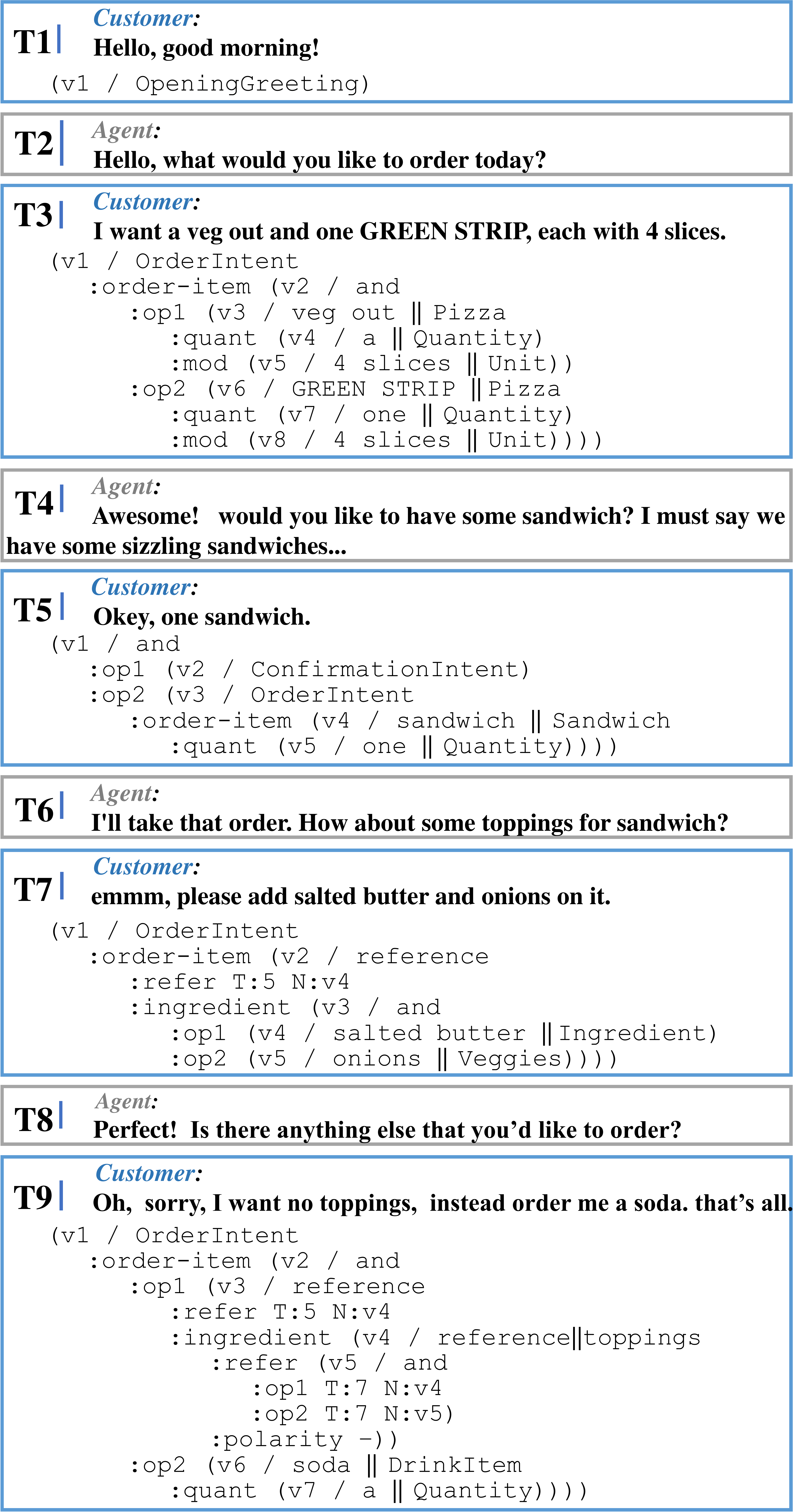}
    \caption{%
        A dialogue example taken from fast food domain. The customer turns are annotated with DMRs.
    }
    \label{fig:sec2}
\end{figure}

\paragraph{Modification}
refers to the semantics where a specific adjective modifies some entities. Modifiers, such as the size or color of an object, are non-essential descriptive content compared to regular arguments~\cite{dowty1982grammatical, dowty1989semantic, dowty2003dual}. The modification semantics is expressed by  \texttt{mod}, for example:

\begin{lstlisting}[escapeinside={(*}{*)},mathescape=true]
(*\textit{\textrm{Veg Out, large}}*)
(Veg Out $\Vert$ Pizza
  :mod (large $\Vert$ Size))
\end{lstlisting}


\paragraph{Quantification}
is also a common semantics in TOD.  Quantification is expressed by the edge labeled with \texttt{quant}. T3 in {\figurename~\ref{fig:sec2}} shows such case similar to the following example:  
\begin{lstlisting}[escapeinside={(*}{*)},mathescape=true]
(*\textit{\textrm{2 burgers}}*)
(burgers $\Vert$ Burger
   :quant (2 $\Vert$ Quantity))
\end{lstlisting}

\paragraph{Conjunction}
refers to the semantic construction that connects elements, e.g., \textit{``A and B''}.  Inspired by AMR, DMR resolves  conjunction with \texttt{Operator} node, such as \texttt{and}, \texttt{or}. Conjunction is important when  multiple intentions  or cumulative entities are expressed. A nested conjunction is like follows:
\begin{lstlisting}[escapeinside={(*}{*)},mathescape=true]
(*\textit{\textrm{sandwich and soda, thank you.}}*)
(and
    :op1 (OrderIntent
            :order-item (and
                            :op1 (sandwich $\Vert$ Sandwich)
                            :op2 (soda $\Vert$ DrinkItem)))
    :op2 ThankYouIntent)
\end{lstlisting}

\paragraph{Cross-Turn Coreference}
is a common phenomenon in dialogues. Since DMR represents semantics in a graph, the implementation of coreference in DMR is to link corresponding nodes, rather than simple text mentions. DMR introduces a special operator {\texttt{Reference}} and edge {\texttt{refer}} to represent cross-turn coreferences. The reference node is with form ``reference $\Vert$ \texttt{<lexical\_value>}'', along with an argument \texttt{refer} that points to another node. For example, T7 in Figure~\ref{fig:sec2} contains a reference node ``v2'' that points to ``T:5 N:v4'', which means this node refers to the node ``v4'' in the T5 turn's DMR.



\paragraph{Negation}
is the  construction which ties a negative polarity to another element, reversing the state of an affair or discontinuing an act.  For instance, the utterance \textit{``Please cancel the burger''} conveys a cancel action to an order. Inspired by AMR, we notice that negation can be seen as a binding act attached to an element. Therefore, instead of representing negation via additional {\texttt{Intent}}, DMR resolves negation by  edge \texttt{polarity} and keyword ``-''. For example, T9 in Figure~\ref{fig:sec2}, node ``v4'' is negated. Moreover, one tricky issue about negation is its scope. A negation act to an order item can be confused as one to an (enclosing) order intent, leading to an unintended ``overkill''. At this stage of development, we make a simplification and restrict the negation to attach only to {\texttt{Entity}}.

\section{Related Work}
\label{sec:related_work}

There is a rising interest in developing more flexible representation for TOD other than the slots representation~\citep{bobrow:1977}. In this section, we briefly introduce them and compare the most related works with DMR.

\paragraph{AMR and Dialogue-AMR} Using AMR for semantic parsing has been studied from a very early time~\cite{DBLP:conf/acllaw/BanarescuBCGGHK13}. 
There are several works that apply AMR to dialogue systems. \citet{DBLP:conf/acl/BaiCS020} model dialogue state with AMR for chit-chat. \citet{bonial-etal-2019-augmenting} extended AMR for human-robot dialogues, and further formalize it as Dialogue-AMR~\citep{bonial-etal-2020-dialogue, bonial-etal-2021-builder}. Dialogue-AMR represents both the illocutionary force and the propositional content of the utterance. Compared to these works, DMR focuses on TOD specifically with extended node types for  TOD description. DMR is intent-centric, and only captures semantics defined in the ontology of the intents. Further, the design of inheritance hierarchy aims at a  better domain generalization to support a broad range of applications. AMR and Dialogue-AMR are closely related to DMR, so we have more detailed comparisons in Appendix~\ref{appendix:comparison} to show the differences between them.

\paragraph{Compositional Intent-Slot} Some recent works focus on the compositional  intent-slot framework, such as TOP~\citep{DBLP:conf/emnlp/GuptaSMKL18}, SB-TOP~\citep{DBLP:conf/emnlp/AghajanyanMSDHL20} and TreeDST~\citep{DBLP:conf/emnlp/ChengAABDFKKLPW20}. These formalizations are much more powerful than the flat intent-slot schema. Generally, they focus less on how to get representations for different domains and give fewer descriptions on how nodes/edges are connected. On the contrary, the  design which contains both domain-agnostic and domain-specific parts allows DMR to be applied and extended to different domains while assuring the maintenance of the representation   structure in the meantime. From this point of view, DMR is designed to provide services to different businesses. This different focus of the application scenarios marks the key difference between DMR and these compositional intent-slot representations.

\paragraph{Programs} Many  efforts have been devoted to explore the  representation in  programs~\citep{DBLP:conf/naacl/Price90,DBLP:conf/aaai/ZelleM96,DBLP:conf/uai/ZettlemoyerC05, DBLP:journals/coling/LiangJK13}. Though powerful expressiveness, they are hard in annotation which makes it limited in proposing large-scale  dialogue datasets. Recently, some works such as SMCalFlow~\citep{SMDataflow2020} and TOC~\citep{campagna2021skim} and ThingTalk~\citep{DBLP:journals/corr/abs-2203-12751, campagna-etal-2022-shot} proposes to use executable dialogue states for TOD. To this end, the representation itself is also a specially-designed programming language. While executing, both database operations and response generation are performed by the program at the same time. DMR keeps the dialogue state architecture, and leaves the implementation of business logic to the user applications.

\section{Data}

We use the fast food domain data from the MultiDoGO dataset proposed by \citet{peskov-etal-2019-multi}. We annotate all the customers' utterances with the redefined ontology. We call the annotated dataset \textbf{DMR-FastFood}. This dataset contains 7k annotated dialogues, and each dialogue has 18.5 turns on average, which is much more than other datasets. Further, there are 7k references, 16k conjunctions and 557 negations annotated. The annotation process, statistics and comparison with related datasets are described in Appendix~\ref{appendix:data_anno}.

\section{NLU with DMR}
As  NLU tasks of the flat intent-slot representation including \emph{Intent Classification} and \emph{Slot Filling},  NLU tasks under  the DMR framework  are to extract DMR graphs from the customer's utterances.
Given a customer's utterance $x_i$ and the dialogue context $(x_0,\cdots,x_{i-1})$, the NLU tasks are to predict the DMR graph $g_i$. In this section, we introduce NLU tasks with DMR and  proposed models.

\subsection{Tasks}
Though most of the DMRs can be predicted by a semantic parsing model, the turns that have cross-turn coreferences, namely  \emph{referring turns}, are not the case. The reference nodes in the referring turns need to be resolved to link to their referent nodes -- nodes that are assigned with variables -- in DMRs from the dialogue context. Parsing DMRs and resolving coreferences for the referring turns at the same time is not a trivial task. Thus, in this work we split NLU with DMR into two sub-tasks: DMR Parsing and Coreference Resolution.

\paragraph{DMR Parsing}
aims to parse a customer utterance into a DMR graph, \emph{without} resolving the reference nodes. This semantic parsing task is similar to the NLU task in most related works, including TOP, SB-TOP, TreeDST and SMCalFlow.

\paragraph{Coreference Resolution}
resolves the reference nodes predicted by DMR parsing. Differing from traditional text-based coreference resolution, which links referring expressions to their antecedents in texts, this task is defined as follows: for each reference node $n_r$ in a referring turn's DMR $g_t$, the task is to predict whether $n_r$ and a given candidate node $n_c \in \{g_0, \cdots, g_{t-1}\}$  are coreferred.

\subsection{Models}

Our overall framework is composed of two stages. One is to parse graphs from the utterance by a Seq2Seq model, then is  to resolve coreferences based on   a GNN model.

\subsubsection{DMR Parsing Model}

The conventional approach to semantic parsing task is the Seq2Seq architecture which inputs the utterances and outputs the linearized tree or graph. This approach is also applied by SMCalFlow, SB-TOP, and \citet{10.1145/3366423.3380064}. We utilize Seq2Seq architecture for DMR parsing as well, and restrict the decoder vocabulary to get more reasonable results. The details of our Seq2Seq model are described as follows:

\paragraph{Input}
We concatenate the utterances to be parsed and the dialogue context -- which are the customer's and agent's utterances in previous turns -- to form the model input according to their order. Specifically,  the model takes the input sequence $r_{i-c}||x_{i-c}, \cdots, r_j||x_j,\cdots, r_{i-1}||x_{i-1}, r_{i}||x_{i}$, where $x_i$ is the customer's utterance to be parsed; $x_j$ is the customer's or agent's utterance in the dialogue context; $c$ is the context size. $r_j$ is the role tag which can be = \texttt{customer:} or \texttt{agent:}, for the customer's and the agent's turn, respectively.

\paragraph{Output}
The output sequence of turn $i$ is the linearized form of $g_i$. To linearize  $g_i$ takes three steps: 1) remove the \texttt{refer} edges of each reference node since the coreference resolution model will resolve them (see Section \ref{gnncoref_desc}), 2) remove the variables of nodes and assign them back in the post-processing step for resolving references, and 3) convert the graph to a bracket expression. For example, the following DMR:
\begin{lstlisting}
(v1 / OrderIntent
  :order-item 
      (v2 / reference
          :refer (T:3 N:v1)
          :mod (v3 / large || Size)))
\end{lstlisting}
is converted to
\begin{lstlisting}
( OrderIntent ( :order-item ( reference ( :mod ( large || Size ) ) ) )
\end{lstlisting}

Since the \texttt{<lexical\_value>} units are from the utterance contents, we constrain the decoding step to only generate tokens from either the schema or the utterance $x_i$. In our model, we mask the probabilities of non-relevant tokens to zeros in the output distribution at each decoding step. The output sequence is then parsed to DMRs with a shift-reduce parser, and the nodes are assigned with variables with Depth-First-Search.

\paragraph{Post-processing}
Though we restrict the decoder vocabulary, it does not guarantee the predicted sequence could be parsed to a valid graph because the sequence could be an invalid bracket expression. To tackle this issue, a rather flexible shift-reduce parser is applied   to get a valid bracket expression by adding missing brackets or removing redundant brackets. If this rescue fails, the prediction is set to \texttt{OutOfDomainIntent}. And then we assign variables to the nodes in the recovered DMR graphs.

\subsubsection{Coreference Resolution Model}
\label{gnncoref_desc}

\begin{figure}[h]
    \centering
    \includegraphics[width=0.9\columnwidth]{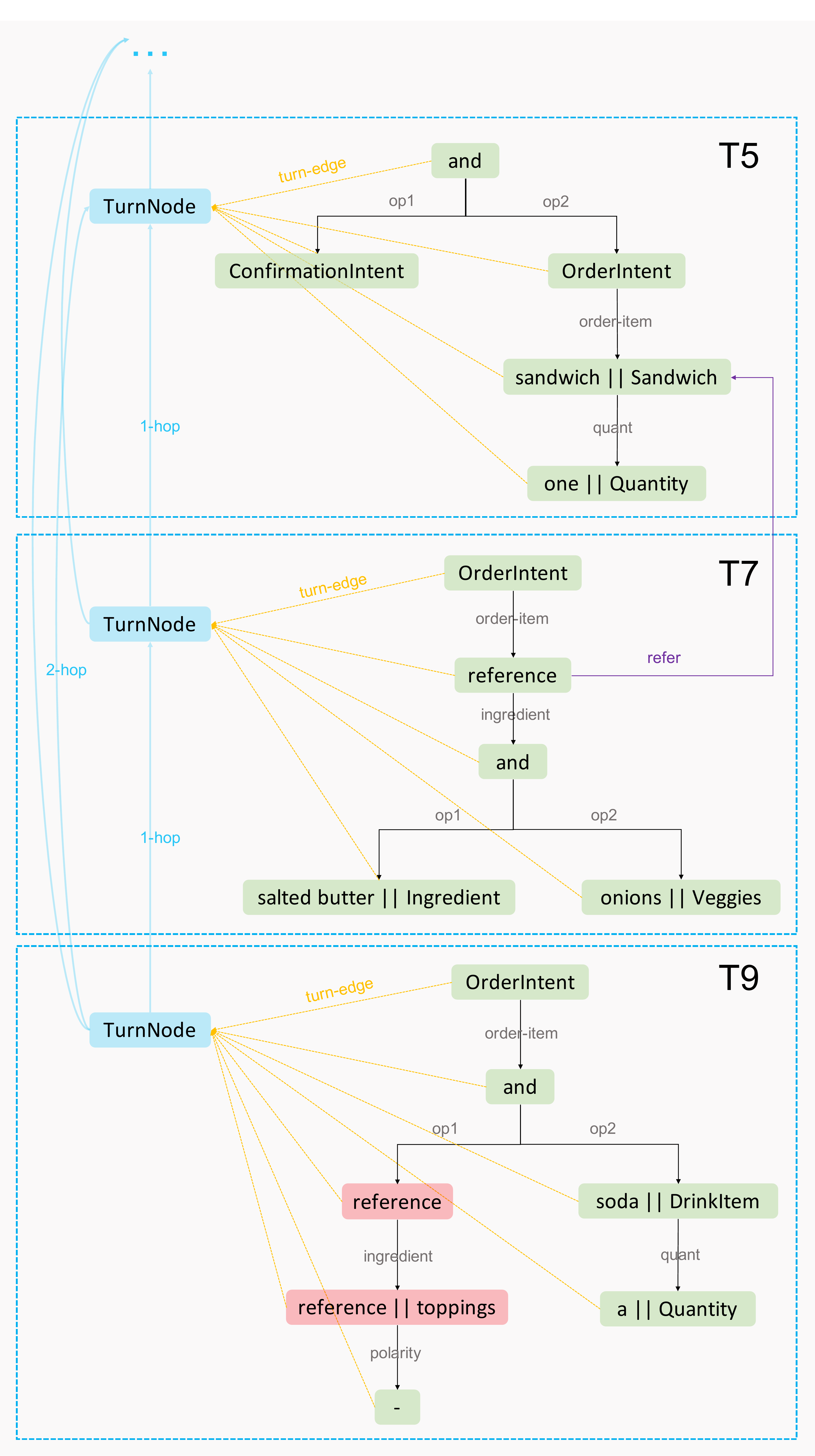}
    \caption{Dialogue Graph for GNNCoref model. The example used here is for resolving reference nodes (pink-colored nodes) in T9. The black arrows are edges within DMRs; the orange arrows are edges from DMR nodes to the turn nodes; the blue arrows are inter-turn edges that link DMRs through turn nodes; and the purple arrow is the edge for linking resolved coreferences in the context.}
    \label{fig:dialogue_graph}
\end{figure}
For the graph-based coreference resolution task, 
we propose a GNN-based model \textbf{GNNCoref}. The following equations show how the model works:

\begin{align}
     & G_t = \text{BuildGraph}(g_0, \cdots, g_t)                                    \\
     & \mathbf{G}_t = \text{GNNEncoder}(G_t)                                        \\
     & p(\text{corefer} | n_r, n_c) = \text{Classifier}(\mathbf{n}_r, \mathbf{n}_c)
\end{align}

First, for each referring turn $t$, a \emph{Dialogue Graph} $G_t$ is built; then the dialogue graph is encoded by a GNN encoder to encode the node features, and the encoded graph is denoted as $\mathbf{G}_t$; $\mathbf{n}_r$ and $\mathbf{n}_c$ in $\mathbf{G}_t$ are the encoded node features of the reference node and candidate node respectively, they are input into a binary classifier to predict whether they are coreferred. Next, we describe the details of these three modules.

\paragraph{Build Dialogue Graph}
The dialogue graph $G_t$ is built by connecting the DMRs $(g_0, \cdots, g_j, \cdots, g_t)$ according to their order in the dialogue. Figure~\ref{fig:dialogue_graph} shows the dialogue graph for resolving references in T9 in Figure~\ref{fig:sec2}. Specifically, we first build a \emph{Turn Graph} for each turn with its DMR graph structure, and link each node to a turn-level global node which we call \emph{turn node} with edge 
labeled \texttt{turn-edge}. The turn graphs are included in the dashed boxes in the figure. Then each turn node (e.g. for $g_j$)   points to its $k$-hop ancestor ($g_{j-k}$) with the edge labeled \texttt{$k$-hop}. These inter-turn edges connect the DMRs to form one connected dialogue graph. Finally, if there are reference nodes already resolved  in the dialogue context, e.g. the reference node in T7 in the figure, they are connected to their referent nodes with edge \texttt{refer}. In this way, the coreference resolution for the current referring turn depends on the previously resolved references, which brings more information for the task. Additionally, we add inverse edges for each edge to allow the message to pass bidirectionally. In dialogue graphs, every two turn graphs are linked through the turn nodes. Since the turn node links to all the nodes in that turn, every node in the dialogue graph can 
connect to each other after a three-step message passing so that  all the context information can be encoded to the nodes.

\paragraph{GNN Encoder}
The graph is encoded with a 3-layer Relational Graph Convolutional Network (R-GCN) \cite{schlichtkrull2018modeling} to encode the edge information to the nodes. The GNN encoder is designed to enhance the message passing among nodes and edges so that global context information can be captured in this process.

\paragraph{Classifier}
The binary classifier is a Multilayer Perceptron (MLP) with a Sigmoid activation for output. In the inference stage, we set a threshold $\beta$ to determine the predictions. In our experiments, the value of $\beta$ is tuned on the development set (see Appendix~\ref{coref_hyp} for details).

For a given reference node $n_r$, treating all the nodes in the context as its candidates is unwise, because $n_r$ has the same entity type as its referents. However, the reference nodes are not labeled with types. According to our annotation guideline for the DMR-FastFood dataset described in Appendix~\ref{refer_annotation}, all the reference nodes have the same incoming edge as the referents, thus, we choose the nodes in the context with the same incoming edge (or have the same incoming edge if there are more than one incoming edges) as $n_r$ to be its candidates.

\section{Experiments}

We report experiments for DMR Parsing and Coreference Resolution separately, and the combined results on the complete DMR graphs. Further, we analyze the key factors that affect the model performance for the two tasks. The hyperparameters used and training details are listed in Appendix~\ref{appendix:exp}.

\subsection{DMR Parsing}
The details of the DMR parsing models are described as follows:

\paragraph{BiLSTM+GloVe}
The encoder is a two-layer bi-directional LSTM (BiLSTM) and the decoder is a two-layer uni-directional LSTM. The word embeddings are initialized with GloVe840B-300d. This model contains 23M parameters.

\paragraph{RoBERTa-base}
The encoder is RoBERTa-base \cite{liu2019roberta}, and the decoder is a two-layer randomly initialized transformer with four attention heads and the same hidden size as the encoder. This model contains 183M parameters.

\paragraph{BART-base}
BART \cite{lewis-etal-2020-bart} is a powerful pretrained encoder-decoder model for Seq2Seq tasks. We finetune the BART-base, directly for DMR Parsing. This model contains 140M parameters.

We use \textbf{Exact Match} accuracy to measure  between predicted DMRs and the ground truths to evaluate the DMR Parsing results. To match the graphs semantically, we utilize the Smatch metric~\cite{cai-knight-2013-smatch} designed for evaluating AMRs.\footnote{The Smatch code we use is adapted from \url{https://github.com/snowblink14/smatch}} Two DMRs are exactly matched if their Smatch score equals 1. Setting context size $c=1$, the performances of the DMR Parsing models are shown in Table~\ref{tab:parsing_results}. The best results are achieved by BART-base model which are more than 10 points over the other two models, showing a well pretrained Seq2Seq model is essential for this task. 


\begin{table}[]
\begin{subtable}{1\textwidth}
\centering
    \begin{tabular}{lccc}
    \toprule
        Model & Dev set & Test set  \\
        \hline
        BiLSTM+GloVe & 65.57 & 65.75 \\
        RoBERTa-base & 69.24 & 68.23  \\
        BART-base & \textbf{82.56} & \textbf{83.39}  \\
    \bottomrule
    \end{tabular}
    \caption{DMR Parsing exact match accuracy ($c=1$).}
    \label{tab:parsing_results}
\end{subtable}

\bigskip
\begin{subtable}{1\textwidth}
\centering
    \begin{tabular}{lccc}
    \toprule
        Method & Train set & Dev set & Test set \\
        \hline
        Rule & 21.03 & 22.77 & 21.19 \\
        MLP & - & 69.92 & 70.42 \\
        GNNCoref & - & \textbf{78.23} & \textbf{79.01} \\
    \bottomrule
    \end{tabular}
    \caption{Coreference Resolution accuracy.}
    \label{tab:coref_results}
\end{subtable}

\bigskip
\begin{subtable}{1\textwidth}
\centering
    \begin{tabular}{lccc}
    \toprule
        Model & Dev set & Test set  \\
        \hline
        BiLSTM+GloVe & 64.09 & 64.30 \\
        RoBERTa-base & 67.57 & 66.69  \\
        BART-base & \textbf{80.73} & \textbf{81.47}  \\
    \bottomrule
    \end{tabular}
    \caption{The overall NLU results. The results are Exact Match of the completed DMR graphs whose reference nodes have been resolved by GNNCoref.}
    \label{tab:overall_results}
\end{subtable}

\caption{Coreference resolution and the overall NLU performances on the DMR-FastFood dataset.}
\label{tab:nlu_results}
\end{table}

\subsection{Coreference Resolution}
\label{coref_results}

To show the effectiveness of the proposed Coreference Resolution model, we compare the results with a heuristic \textbf{rule-based} method and a \textbf{MLP} baseline model. The rule-based method selects the last DMR graph in the context and selects the candidate nodes in this DMR as the predicted referents. This distance-based heuristic is commonly used as an important feature in coreference resolution \citep{bengtson-roth-2008-understanding}. In the MLP model, the features of the reference node and candidate nodes are the average of the word embeddings of their one-hop neighbor in the DMR graph and their own, and the features of the reference node and candidate node are concatenated to input into a 2-layer MLP classifier. For GNNCoref model, the initial node features for entity nodes and reference nodes are the average of GloVe6B-100d embeddings~ \cite{pennington2014glove} of all tokens (
except for the variable) in the node. Other nodes are symbols defined in the DMR-FastFood ontology and their embeddings are randomly initialized. In our experiments, we use the DGL~\cite{wang2019dgl} implementation of R-GCN.
All the methods are trained and evaluated on ground truth DMRs.

We measure coreference resolution with accuracy, i.e., a reference node is resolved correctly if the predicted turns and nodes are the same as the ground truth. Note that about 31.2\% of the reference nodes in DMR-FastFood dataset have only one candidate which is directly their referent, we ignore these cases during training and evaluation. The results are listed in Table~\ref{tab:coref_results}. We can see that a simple heuristic rule can't handle this task well. Also, GNNCoref outperforms MLP well indicating the global dialogue context information captured with the graph structure is very useful compared to the local one-hop features.

\subsection{The Overall NLU Performance}

Combining the predictions of the DMR Parsing and Coreference Resolution model, we get the complete DMR graphs. The exact match of the complete DMRs are shown in Table~\ref{tab:overall_results}, the coreference resolution predictions used here are by GNNCoref reported in Table~\ref{tab:coref_results}. Comparing to the parsing results in Table~\ref{tab:parsing_results}, the performance drops less than two points which proves the effectiveness of the two-step approach to this NLU task.

\subsection{Error Analysis of DMR Parsing}

We conduct error analysis to explore the difficulties and room for improvement of the DMR Parsing task. First, we analyze  four types of errors: 
\begin{itemize}
    \item \textbf{Invalid Graph}, denotes that the predicted sequence cannot be parsed into a DMR graph with the shift-reduce parser. It is similar to Tree Validity used in \citet{DBLP:conf/emnlp/GuptaSMKL18}.
    \item \textbf{Ontology Mismatch}, denotes that parts of the graph structure are not aligned with the definition in the ontology, e.g. edge \texttt{:order-item} points to a quantity, a \texttt{Pizza} entity with argument \texttt{:address}, etc.
    \item \textbf{Wrong Intent}, denotes wrong intents prediction. As intents are the first-class citizen which would directly affect the behavior of the chatbot agent. We consider a predicted DMR graph with wrong intents when the set of intents in it are not exactly matched with the golden DMR.
    \item \textbf{Compositional Errors}, denotes wrong or missing compositions in complex structures. In DMR-FastFood, we only care about the errors for \texttt{OrderIntent}. We extract the \texttt{OrderIntent} sub-graph from the golden DMR and the predicted DMR, and consider it as a compositional error if their Smatch score is not 1. 
\end{itemize}

We take the error cases from the parsing results of BART-base model shown in Table~\ref{tab:parsing_results} to conduct the analysis. Table~\ref{tab:parsing_error_analysis_1} shows the portions of the errors in the development and test sets respectively. We can see that the model can mostly generate valid graphs which also match the ontology. The main errors are compositional errors and wrong intent, and we observe that 79\% of the wrong intent cases are the utterances with multiple intents. These results indicate that compositional generalization is the main bottleneck of Seq2Seq parsers.

\begin{table}[]
    \centering
    \begin{tabular}{lcc}
    \toprule
      Error Type   &  Dev set & Test set \\
      \hline
      Invalid Graph & 3.7\% & 2.9\% \\
      Ontology Mismatch & 2.3\% & 0.8\% \\
      Wrong Intent & 31.1\% & 39.7\% \\
      Compositional Error & 54.1\% & 46.0\% \\ 
    \bottomrule
    \end{tabular}
    \caption{Portions of errors in the error cases of BART-base model with $c=1$.}
    \label{tab:parsing_error_analysis_1}
\end{table}

Second, we analyze which factors affects the performance of the DMR parsing models. The details are listed in Appendix~\ref{appendix:analyze_parsing}. The main conclusions are intuitive: 1) DMR parsing is dependent on the dialogue context, and 2) longer utterances, deeper and larger DMR graphs make the parsing task harder. 

\subsection{Ablation Study of GNNCoref model}

We conduct ablation studies to investigate the effectiveness of the two key designs in the dialogue graph for GNNCoref: 1) the  global nodes \emph{Turn Node} connecting DMRs through turn-level, and 2)   depending on resolved coreferences in the context by adding \texttt{refer} edges  as described in Section \ref{gnncoref_desc}. We remove the turn nodes by connecting the DMRs through their root nodes instead, and remove the dependence on the resolved coreferences by removing \texttt{refer} edges. As shown in Table~\ref{tab:coref_ablation}, both of the  results are declined compared to original GNNCoref.  However,  without the help of turn node as a global node, performance drops  sharply (From 79.01 to 70.56 on Test set), which indicates that the turn nodes are  more critical.

In order to more rigorously prove the importance of the dialogue context information, we further conduct experiments of   fewer R-GCN layers for GNNCoref. The results are shown in the last two rows of the table. Theoretically, With a 2-layer R-GCN, the nodes can only see information within their turns, thus, no dialogue context information is captured; the 1-layer R-GCN only captures one-hop information for each node. We can see that the performances declined which indicates that less  captured context  leads to lower  performance.
\begin{table}[]
    \centering
    \begin{tabular}{lcc}
        \toprule
                                            & Dev set        & Test set       \\
        \hline
        GNNCoref                            & \textbf{78.23} & \textbf{79.01} \\
        - \textit{Turn Node}                & 68.45          & 70.56          \\
        - \textit{Depend on Resolved Coref} & 76.93          & 78.02          \\
        2-layer R-GCN                       & 76.75          & 76.19          \\
        1-layer R-GCN                       & 50.73          & 47.73              \\
        \bottomrule
    \end{tabular}

    \caption{Ablation study of GNNCoref.}
    \label{tab:coref_ablation}
\end{table}

\section{Conclusion}
In this paper, we focus on the representation with the expression ability of both complex compositional semantics and task-oriented semantics and propose DMR which is capable of representing complex linguistic constructions with a high transferability across domains. 
Moreover, we design the inheritance hierarchy which allows reusing, extending and inheriting node types to enable DMR scale to different domains easily. We annotated a large dataset on fast-food ordering domain, named DMR-FastFood, to incentive research on semantics parsing, which contains more than 70k  utterances annotated with rich linguistic semantics.  We conduct experiments on DMR parsing and  coreference resolution tasks.  Experimental results show that   pre-trained Seq2Seq models could improve the DMR parsing results. We also propose a graph-based model, GNNCoref, for the coreference resolution on DMRs.

\section*{Limitations}
In this work, we propose DMR and a dialogue dataset on the fast-food domain annotated with it. There are some limitations of this work we describe as follows. Firstly, DMR is designed to support a broad range of domains and applications for task-oriented dialogue. However,  because of the human resources, and the observation that enough compositional semantics to begin with, such as conjunction, modification, and negation are contained in  fast-food domain data, our dataset is only annotated on the fast-food domain for now. Secondly, we use DMR for natural language understanding task in TOD. In the real-world TOD systems, the complete pipeline also includes dialogue state tracking, dialogue policy and response generation that we do not deal with. Thirdly, annotating conversations with DMR is more expensive than annotating with intents and slots. Few-shot learning and transfer learning for DMR parsing and coreference resolution could address this issue, and we leave them to future work. And Lastly, compared with the transition-based parser, the Seq2Seq-based semantic parser is not guaranteed to generate well-formed DMRs as it's not introduced with the inductive bias of the ontology, and it needs more data to train.

 



\bibliography{anthology,custom}

\begin{thebibliography}{31}
\expandafter\ifx\csname natexlab\endcsname\relax\def\natexlab#1{#1}\fi

\bibitem[{Aghajanyan et~al.(2020)Aghajanyan, Maillard, Shrivastava, Diedrick,
  Haeger, Li, Mehdad, Stoyanov, Kumar, Lewis, and
  Gupta}]{DBLP:conf/emnlp/AghajanyanMSDHL20}
Armen Aghajanyan, Jean Maillard, Akshat Shrivastava, Keith Diedrick, Michael
  Haeger, Haoran Li, Yashar Mehdad, Veselin Stoyanov, Anuj Kumar, Mike Lewis,
  and Sonal Gupta. 2020.
\newblock \href {https://doi.org/10.18653/v1/2020.emnlp-main.408}
  {Conversational semantic parsing}.
\newblock In \emph{Proceedings of the 2020 Conference on Empirical Methods in
  Natural Language Processing (EMNLP)}, pages 5026--5035, Online. Association
  for Computational Linguistics.

\bibitem[{Bai et~al.(2021)Bai, Chen, Song, and Zhang}]{DBLP:conf/acl/BaiCS020}
Xuefeng Bai, Yulong Chen, Linfeng Song, and Yue Zhang. 2021.
\newblock \href {https://doi.org/10.18653/v1/2021.acl-long.342} {Semantic
  representation for dialogue modeling}.
\newblock In \emph{Proceedings of the 59th Annual Meeting of the Association
  for Computational Linguistics and the 11th International Joint Conference on
  Natural Language Processing (Volume 1: Long Papers)}, pages 4430--4445,
  Online. Association for Computational Linguistics.

\bibitem[{Banarescu et~al.(2013)Banarescu, Bonial, Cai, Georgescu, Griffitt,
  Hermjakob, Knight, Koehn, Palmer, and
  Schneider}]{DBLP:conf/acllaw/BanarescuBCGGHK13}
Laura Banarescu, Claire Bonial, Shu Cai, Madalina Georgescu, Kira Griffitt, Ulf
  Hermjakob, Kevin Knight, Philipp Koehn, Martha Palmer, and Nathan Schneider.
  2013.
\newblock \href {https://aclanthology.org/W13-2322} {{A}bstract {M}eaning
  {R}epresentation for sembanking}.
\newblock In \emph{Proceedings of the 7th Linguistic Annotation Workshop and
  Interoperability with Discourse}, pages 178--186, Sofia, Bulgaria.
  Association for Computational Linguistics.

\bibitem[{Bengtson and Roth(2008)}]{bengtson-roth-2008-understanding}
Eric Bengtson and Dan Roth. 2008.
\newblock \href {https://aclanthology.org/D08-1031} {Understanding the value of
  features for coreference resolution}.
\newblock In \emph{Proceedings of the 2008 Conference on Empirical Methods in
  Natural Language Processing}, pages 294--303, Honolulu, Hawaii. Association
  for Computational Linguistics.

\bibitem[{Bobrow et~al.(1977)Bobrow, Kaplan, Kay, Norman, Thompson, and
  Winograd}]{bobrow:1977}
Daniel~G. Bobrow, Ronald~M. Kaplan, Martin Kay, Donald~A. Norman, Henry~S.
  Thompson, and Terry Winograd. 1977.
\newblock \href {https://doi.org/10.1016/0004-3702(77)90018-2} {Gus, {A}
  frame-driven dialog system}.
\newblock \emph{Artif. Intell.}, 8(2):155--173.

\bibitem[{Bonial et~al.(2021)Bonial, Abrams, Traum, and
  Voss}]{bonial-etal-2021-builder}
Claire Bonial, Mitchell Abrams, David Traum, and Clare Voss. 2021.
\newblock \href {https://aclanthology.org/2021.iwcs-1.17} {Builder, we have
  done it: Evaluating {\&} extending dialogue-{AMR} {NLU} pipeline for two
  collaborative domains}.
\newblock In \emph{Proceedings of the 14th International Conference on
  Computational Semantics (IWCS)}, pages 173--183, Groningen, The Netherlands
  (online). Association for Computational Linguistics.

\bibitem[{Bonial et~al.(2020)Bonial, Donatelli, Abrams, Lukin, Tratz, Marge,
  Artstein, Traum, and Voss}]{bonial-etal-2020-dialogue}
Claire Bonial, Lucia Donatelli, Mitchell Abrams, Stephanie~M. Lukin, Stephen
  Tratz, Matthew Marge, Ron Artstein, David Traum, and Clare Voss. 2020.
\newblock \href {https://aclanthology.org/2020.lrec-1.86} {Dialogue-{AMR}:
  {A}bstract {M}eaning {R}epresentation for dialogue}.
\newblock In \emph{Proceedings of the 12th Language Resources and Evaluation
  Conference}, pages 684--695, Marseille, France. European Language Resources
  Association.

\bibitem[{Bonial et~al.(2019)Bonial, Donatelli, Lukin, Tratz, Artstein, Traum,
  and Voss}]{bonial-etal-2019-augmenting}
Claire Bonial, Lucia Donatelli, Stephanie~M. Lukin, Stephen Tratz, Ron
  Artstein, David Traum, and Clare Voss. 2019.
\newblock \href {https://doi.org/10.18653/v1/W19-3322} {Augmenting {A}bstract
  {M}eaning {R}epresentation for human-robot dialogue}.
\newblock In \emph{Proceedings of the First International Workshop on Designing
  Meaning Representations}, pages 199--210, Florence, Italy. Association for
  Computational Linguistics.

\bibitem[{Cai and Knight(2013)}]{cai-knight-2013-smatch}
Shu Cai and Kevin Knight. 2013.
\newblock \href {https://aclanthology.org/P13-2131} {{S}match: an evaluation
  metric for semantic feature structures}.
\newblock In \emph{Proceedings of the 51st Annual Meeting of the Association
  for Computational Linguistics (Volume 2: Short Papers)}, pages 748--752,
  Sofia, Bulgaria. Association for Computational Linguistics.

\bibitem[{Campagna et~al.(2022)Campagna, Semnani, Kearns, Koba~Sato, Xu, and
  Lam}]{campagna-etal-2022-shot}
Giovanni Campagna, Sina Semnani, Ryan Kearns, Lucas~Jun Koba~Sato, Silei Xu,
  and Monica Lam. 2022.
\newblock \href {https://aclanthology.org/2022.findings-acl.317} {A few-shot
  semantic parser for {W}izard-of-{O}z dialogues with the precise {T}hing{T}alk
  representation}.
\newblock In \emph{Findings of the Association for Computational Linguistics:
  ACL 2022}, pages 4021--4034, Dublin, Ireland. Association for Computational
  Linguistics.

\bibitem[{Campagna et~al.(2021)Campagna, Semnani, Kearns, Sato, Xu, and
  Lam}]{campagna2021skim}
Giovanni Campagna, Sina~J. Semnani, Ryan Kearns, Lucas Jun~Koba Sato, Silei Xu,
  and Monica~S. Lam. 2021.
\newblock \href {http://arxiv.org/abs/2009.07968} {Skim : Few-shot
  conversational semantic parsers with formal dialogue contexts}.

\bibitem[{Cheng et~al.(2020)Cheng, Agrawal, Mart{\'\i}nez~Alonso, Bhargava,
  Driesen, Flego, Kaplan, Kartsaklis, Li, Piraviperumal, Williams, Yu,
  {\'O}~S{\'e}aghdha, and Johannsen}]{DBLP:conf/emnlp/ChengAABDFKKLPW20}
Jianpeng Cheng, Devang Agrawal, H{\'e}ctor Mart{\'\i}nez~Alonso, Shruti
  Bhargava, Joris Driesen, Federico Flego, Dain Kaplan, Dimitri Kartsaklis, Lin
  Li, Dhivya Piraviperumal, Jason~D. Williams, Hong Yu, Diarmuid
  {\'O}~S{\'e}aghdha, and Anders Johannsen. 2020.
\newblock \href {https://doi.org/10.18653/v1/2020.emnlp-main.651}
  {Conversational semantic parsing for dialog state tracking}.
\newblock In \emph{Proceedings of the 2020 Conference on Empirical Methods in
  Natural Language Processing (EMNLP)}, pages 8107--8117, Online. Association
  for Computational Linguistics.

\bibitem[{Dowty(1982)}]{dowty1982grammatical}
David Dowty. 1982.
\newblock Grammatical relations and montague grammar.
\newblock In \emph{The nature of syntactic representation}, pages 79--130.
  Springer.

\bibitem[{Dowty(2003)}]{dowty2003dual}
David Dowty. 2003.
\newblock The dual analysis of adjuncts/complements in categorial grammar.
\newblock \emph{Modifying adjuncts}, 33.

\bibitem[{Dowty(1989)}]{dowty1989semantic}
David~R Dowty. 1989.
\newblock On the semantic content of the notion of ‘thematic role’.
\newblock In \emph{Properties, types and meaning}, pages 69--129. Springer.

\bibitem[{Fillmore(1968)}]{fillmore:1968}
Charles~J. Fillmore. 1968.
\newblock The case for the case.
\newblock In E.~Bach and R.~Harms, editors, \emph{Universals in Linguistic
  Theory}. Rinehart and Winston, New York.

\bibitem[{Fleiss(1971)}]{fleiss1971measuring}
Joseph~L Fleiss. 1971.
\newblock Measuring nominal scale agreement among many raters.
\newblock \emph{Psychological bulletin}, 76(5):378.

\bibitem[{Gupta et~al.(2018)Gupta, Shah, Mohit, Kumar, and
  Lewis}]{DBLP:conf/emnlp/GuptaSMKL18}
Sonal Gupta, Rushin Shah, Mrinal Mohit, Anuj Kumar, and Mike Lewis. 2018.
\newblock \href {https://doi.org/10.18653/v1/D18-1300} {Semantic parsing for
  task oriented dialog using hierarchical representations}.
\newblock In \emph{Proceedings of the 2018 Conference on Empirical Methods in
  Natural Language Processing}, pages 2787--2792, Brussels, Belgium.
  Association for Computational Linguistics.

\bibitem[{Lam et~al.(2022)Lam, Campagna, Moradshahi, Semnani, and
  Xu}]{DBLP:journals/corr/abs-2203-12751}
Monica~S. Lam, Giovanni Campagna, Mehrad Moradshahi, Sina~J. Semnani, and Silei
  Xu. 2022.
\newblock \href {https://doi.org/10.48550/arXiv.2203.12751} {Thingtalk: An
  extensible, executable representation language for task-oriented dialogues}.
\newblock \emph{CoRR}, abs/2203.12751.

\bibitem[{Lewis et~al.(2020)Lewis, Liu, Goyal, Ghazvininejad, Mohamed, Levy,
  Stoyanov, and Zettlemoyer}]{lewis-etal-2020-bart}
Mike Lewis, Yinhan Liu, Naman Goyal, Marjan Ghazvininejad, Abdelrahman Mohamed,
  Omer Levy, Veselin Stoyanov, and Luke Zettlemoyer. 2020.
\newblock \href {https://doi.org/10.18653/v1/2020.acl-main.703} {{BART}:
  Denoising sequence-to-sequence pre-training for natural language generation,
  translation, and comprehension}.
\newblock In \emph{Proceedings of the 58th Annual Meeting of the Association
  for Computational Linguistics}, pages 7871--7880, Online. Association for
  Computational Linguistics.

\bibitem[{Liang et~al.(2011)Liang, Jordan, and
  Klein}]{DBLP:journals/coling/LiangJK13}
Percy Liang, Michael Jordan, and Dan Klein. 2011.
\newblock \href {https://aclanthology.org/P11-1060} {Learning dependency-based
  compositional semantics}.
\newblock In \emph{Proceedings of the 49th Annual Meeting of the Association
  for Computational Linguistics: Human Language Technologies}, pages 590--599,
  Portland, Oregon, USA. Association for Computational Linguistics.

\bibitem[{Liu et~al.(2019)Liu, Ott, Goyal, Du, Joshi, Chen, Levy, Lewis,
  Zettlemoyer, and Stoyanov}]{liu2019roberta}
Yinhan Liu, Myle Ott, Naman Goyal, Jingfei Du, Mandar Joshi, Danqi Chen, Omer
  Levy, Mike Lewis, Luke Zettlemoyer, and Veselin Stoyanov. 2019.
\newblock \href {https://arxiv.org/abs/1907.11692} {Roberta: A robustly
  optimized bert pretraining approach}.
\newblock \emph{arXiv preprint arXiv:1907.11692}.

\bibitem[{Pennington et~al.(2014)Pennington, Socher, and
  Manning}]{pennington2014glove}
Jeffrey Pennington, Richard Socher, and Christopher Manning. 2014.
\newblock \href {https://doi.org/10.3115/v1/D14-1162} {{G}lo{V}e: Global
  vectors for word representation}.
\newblock In \emph{Proceedings of the 2014 Conference on Empirical Methods in
  Natural Language Processing ({EMNLP})}, pages 1532--1543, Doha, Qatar.
  Association for Computational Linguistics.

\bibitem[{Peskov et~al.(2019)Peskov, Clarke, Krone, Fodor, Zhang, Youssef, and
  Diab}]{peskov-etal-2019-multi}
Denis Peskov, Nancy Clarke, Jason Krone, Brigi Fodor, Yi~Zhang, Adel Youssef,
  and Mona Diab. 2019.
\newblock \href {https://doi.org/10.18653/v1/D19-1460} {Multi-domain
  goal-oriented dialogues ({M}ulti{D}o{GO}): Strategies toward curating and
  annotating large scale dialogue data}.
\newblock In \emph{Proceedings of the 2019 Conference on Empirical Methods in
  Natural Language Processing and the 9th International Joint Conference on
  Natural Language Processing (EMNLP-IJCNLP)}, pages 4526--4536, Hong Kong,
  China. Association for Computational Linguistics.

\bibitem[{Price(1990)}]{DBLP:conf/naacl/Price90}
Patti~J. Price. 1990.
\newblock \href {https://aclanthology.org/H90-1020} {Evaluation of spoken
  language systems: the {ATIS} domain}.
\newblock In \emph{Speech and Natural Language: Proceedings of a Workshop Held
  at Hidden Valley, {P}ennsylvania, June 24-27,1990}.

\bibitem[{Rongali et~al.(2020)Rongali, Soldaini, Monti, and
  Hamza}]{10.1145/3366423.3380064}
Subendhu Rongali, Luca Soldaini, Emilio Monti, and Wael Hamza. 2020.
\newblock \href {https://doi.org/10.1145/3366423.3380064} {Don't parse,
  generate! {A} sequence to sequence architecture for task-oriented semantic
  parsing}.
\newblock In \emph{{WWW} '20: The Web Conference 2020, Taipei, Taiwan, April
  20-24, 2020}, pages 2962--2968. {ACM} / {IW3C2}.

\bibitem[{Schlichtkrull et~al.(2018)Schlichtkrull, Kipf, Bloem, Van Den~Berg,
  Titov, and Welling}]{schlichtkrull2018modeling}
Michael Schlichtkrull, Thomas~N Kipf, Peter Bloem, Rianne Van Den~Berg, Ivan
  Titov, and Max Welling. 2018.
\newblock Modeling relational data with graph convolutional networks.
\newblock In \emph{European semantic web conference}, pages 593--607. Springer.

\bibitem[{{Semantic Machines} et~al.(2020){Semantic Machines}, Andreas, Bufe,
  Burkett, Chen, Clausman, Crawford, Crim, DeLoach, Dorner, Eisner, Fang, Guo,
  Hall, Hayes, Hill, Ho, Iwaszuk, Jha, Klein, Krishnamurthy, Lanman, Liang,
  Lin, Lintsbakh, McGovern, Nisnevich, Pauls, Petters, Read, Roth, Roy, Rusak,
  Short, Slomin, Snyder, Striplin, Su, Tellman, Thomson, Vorobev, Witoszko,
  Wolfe, Wray, Zhang, and Zotov}]{SMDataflow2020}
{Semantic Machines}, Jacob Andreas, John Bufe, David Burkett, Charles Chen,
  Josh Clausman, Jean Crawford, Kate Crim, Jordan DeLoach, Leah Dorner, Jason
  Eisner, Hao Fang, Alan Guo, David Hall, Kristin Hayes, Kellie Hill, Diana Ho,
  Wendy Iwaszuk, Smriti Jha, Dan Klein, Jayant Krishnamurthy, Theo Lanman,
  Percy Liang, Christopher~H. Lin, Ilya Lintsbakh, Andy McGovern, Aleksandr
  Nisnevich, Adam Pauls, Dmitrij Petters, Brent Read, Dan Roth, Subhro Roy,
  Jesse Rusak, Beth Short, Div Slomin, Ben Snyder, Stephon Striplin, Yu~Su,
  Zachary Tellman, Sam Thomson, Andrei Vorobev, Izabela Witoszko, Jason Wolfe,
  Abby Wray, Yuchen Zhang, and Alexander Zotov. 2020.
\newblock \href {https://doi.org/10.1162/tacl_a_00333} {Task-oriented dialogue
  as dataflow synthesis}.
\newblock \emph{Transactions of the Association for Computational Linguistics},
  8:556--571.

\bibitem[{Wang et~al.(2019)Wang, Zheng, Ye, Gan, Li, Song, Zhou, Ma, Yu, Gai,
  Xiao, He, Karypis, Li, and Zhang}]{wang2019dgl}
Minjie Wang, Da~Zheng, Zihao Ye, Quan Gan, Mufei Li, Xiang Song, Jinjing Zhou,
  Chao Ma, Lingfan Yu, Yu~Gai, Tianjun Xiao, Tong He, George Karypis, Jinyang
  Li, and Zheng Zhang. 2019.
\newblock \href {https://arxiv.org/abs/1909.01315} {Deep graph library: A
  graph-centric, highly-performant package for graph neural networks}.
\newblock \emph{arXiv preprint arXiv:1909.01315}.

\bibitem[{Zelle and Mooney(1996)}]{DBLP:conf/aaai/ZelleM96}
John~M. Zelle and Raymond~J. Mooney. 1996.
\newblock \href {http://www.aaai.org/Library/AAAI/1996/aaai96-156.php}
  {Learning to parse database queries using inductive logic programming}.
\newblock In \emph{Proceedings of the Thirteenth National Conference on
  Artificial Intelligence and Eighth Innovative Applications of Artificial
  Intelligence Conference, {AAAI} 96, {IAAI} 96, Portland, Oregon, USA, August
  4-8, 1996, Volume 2}, pages 1050--1055. {AAAI} Press / The {MIT} Press.

\bibitem[{Zettlemoyer and Collins(2005)}]{DBLP:conf/uai/ZettlemoyerC05}
Luke~S. Zettlemoyer and Michael Collins. 2005.
\newblock \href
  {https://dslpitt.org/uai/displayArticleDetails.jsp?mmnu=1\&smnu=2\&article\_id=1209\&proceeding\_id=21}
  {Learning to map sentences to logical form: Structured classification with
  probabilistic categorial grammars}.
\newblock In \emph{{UAI} '05, Proceedings of the 21st Conference in Uncertainty
  in Artificial Intelligence, Edinburgh, Scotland, July 26-29, 2005}, pages
  658--666. {AUAI} Press.

\end{thebibliography}
\bibliographystyle{acl_natbib}

\newpage
\clearpage

\appendix

\section{Data Annotation and Statistics}
\label{appendix:data_anno}

The data annotation process has two stages: 1) DMR graph annotation and 2) reference annotation.

\subsection{DMR Annotation}
\begin{figure}[t]
    \centering
    \begin{subfigure}[b]{\textwidth}
        \includegraphics[width=\textwidth]{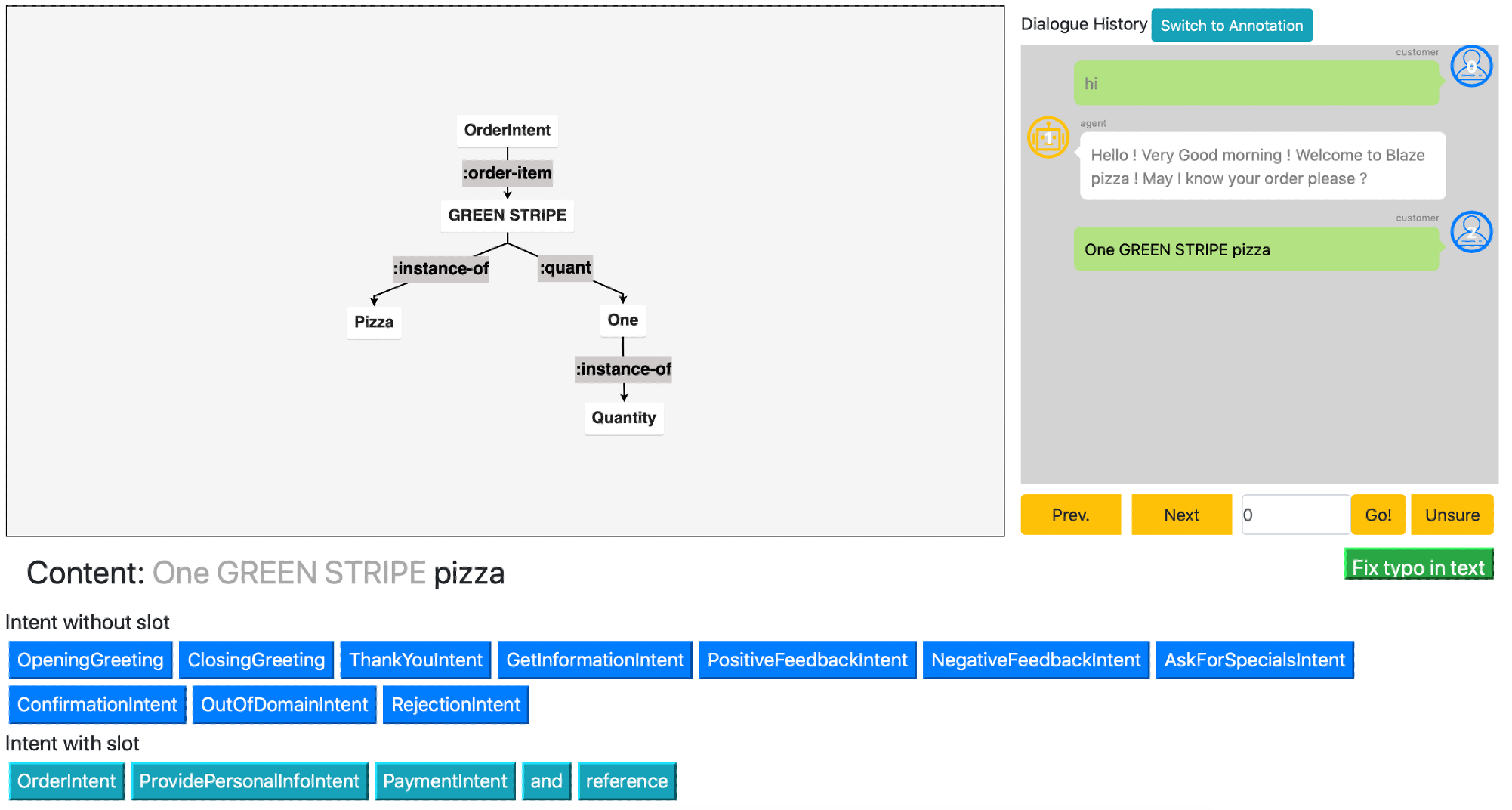}
        \caption{DMR graph annotation tool.}
        \label{fig:dmr_anno_tool}
    \end{subfigure}
    \begin{subfigure}[b]{\textwidth}
        \includegraphics[width=\textwidth]{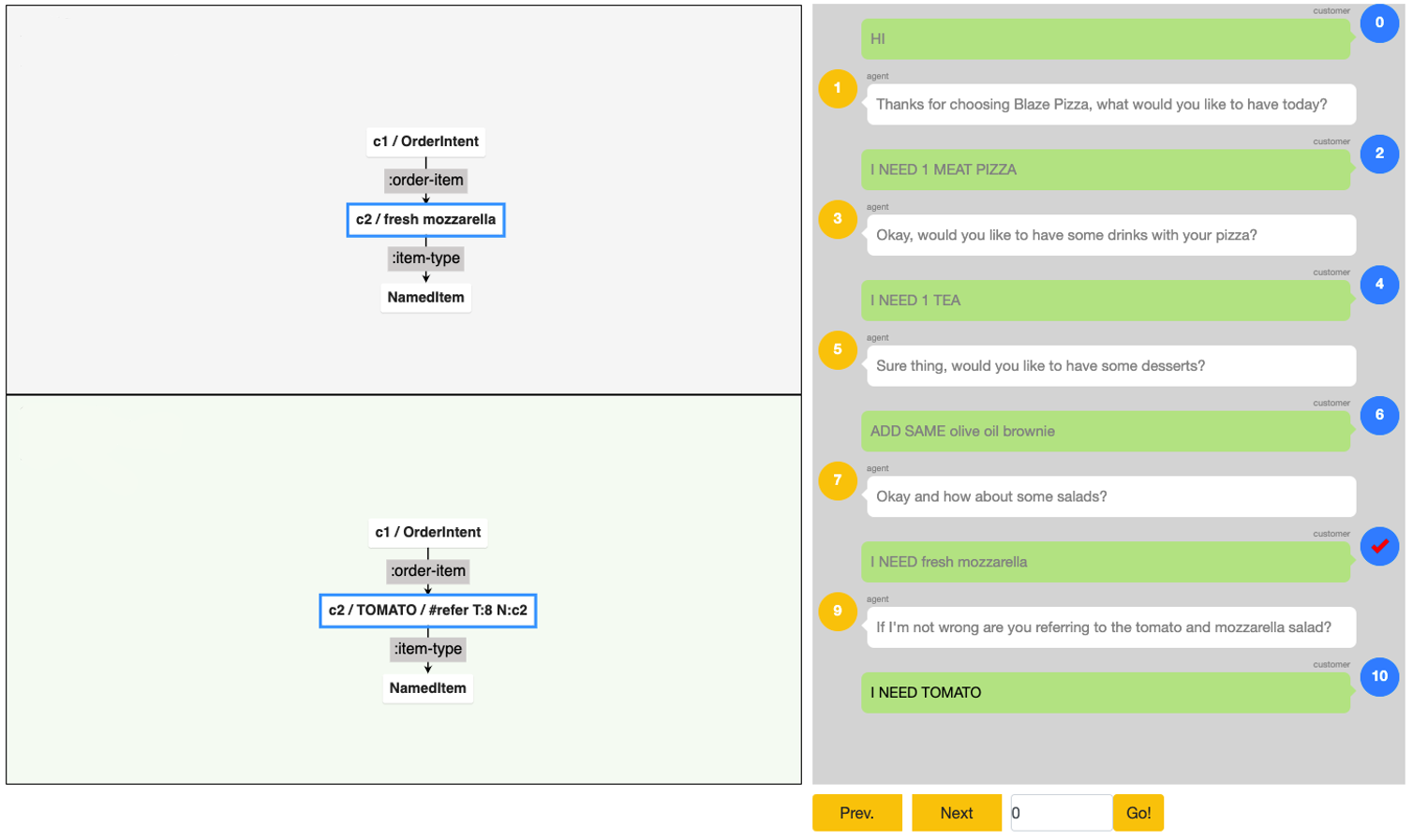}
        \caption{Reference annotation tool.}
        \label{fig:refer_anno_tool}
    \end{subfigure}
    \caption{The interface of the annotation tool.}
    \label{fig:anno_tool}
\end{figure}

In this stage, the annotators draw DMR graphs for each customer turn. They also annotate the referred turn number for the reference nodes. We developed an annotation tool based on GoJS\footnote{\url{https://github.com/NorthwoodsSoftware/GoJS}} for quickly drawing DMR graphs. As shown in Figure~\ref{fig:dmr_anno_tool}, the right part of the tool shows the utterances of dialogue up to the current turn, and the left part is the area for drawing graphs.

Before the annotation, the annotators followed a detailed guideline and took a training process. They draw the DMR graph in the diagram by adding nodes and linking them together. We enable the graph drawing process to follow the schema, which guarantees the validity of the resulting DMR.

To ensure the annotators do not hallucinate node values, the annotators must either select nodes in the bottom, or copy tokens from the current utterance (tokenized by Spacy\footnote{\url{https://spacy.io/}}) to fill the node. Also, there are sanity checks before saving the annotations to the database. After the graph annotation, we assign variables to the nodes in each DMR.

\subsection{Reference Annotation}
\label{refer_annotation}

In this stage, the annotators are given the current turn’s DMR and the referred turn’s DMR, and they need to annotate the referents for each reference node. The tool is modified as Figure~\ref{fig:refer_anno_tool}, and the left part has two diagrams. The below diagram shows the DMR graph for the current turn, and there are reference nodes there. When clicking the reference node, the referred DMR graph will appear in the above diagram, and the annotator can select the referents in it. Further, we constrain the annotator to only select referents with the same incoming edges as the reference nodes.

\subsection{Quality Control}
We make some efforts to ensure the high quality of the dataset.

First, we ask the annotators to fix the typos in the utterances during the annotation process.

Second, in some dialogues, reference nodes appear in the first customer turn, mainly due to the customer ordering toppings but no food items. We remove these dialogues from the data.

The third is double annotation. Though the annotators are well-trained experts, we have 10\% of the dialogues double annotated. We compute Fleiss’ kappa~ \cite{fleiss1971measuring} for measuring the Inter-Annotator Agreement (IAA). After cleaning the annotation, 5,159 utterances have valid double annotations, and the IAA is 0.748, which is a substantial agreement.

\subsection{Data Statistics}

The statistics of DMR-FastFood are listed in Table~\ref{tab:data_stat}, we split the dataset as the original setting.

\begin{table}[h]
    \centering
    \setlength{\tabcolsep}{1mm}
    \begin{tabular}{lccc}
         \toprule
                                   & Train   & Dev    & Test   \\
                \hline
                Dialogue           & 5,585   & 710    & 899    \\
                Utterance          & 102,843 & 13,111 & 16,889 \\
                Utterance/Dialogue & 18.41   & 18.47  & 18.79  \\
                Customer Utterance & 54,465  & 6,911  & 8,952  \\
                Utterance for NLU  & 23,633  & 4,256  & 5,581  \\
                Utterance Length   & 10.24   & 10.28  & 10.25  \\
                Reference          & 6,007   & 802    & 1,039  \\
                Negation           & 430     & 62     & 65     \\
                Conjunction        & 11,770  & 1,499  & 1,989  \\
                NLU DMR Depth      & 2.43    & 2.66   & 2.64   \\
                NLU DMR Nodes      & 3.18    & 3.46   & 3.43   \\
                \bottomrule
    \end{tabular}
    \caption{Detailed statistics of DMR-FastFood dataset.}
    \label{tab:data_stat}
\end{table}

 We add marks and exclude the following utterances: first, we omit utterance-annotation pairs in the train set that also occur in the dev and test set, for including them will cause information leakage; and second, a portion of the data  annotated with a single intent are  excluded  from the dev and test set, since they are more like text classification and  trivial to get right. The left data is used for NLU. In Table~\ref{tab:data_stat}, ``Utterance for NLU'', ``NLU DMR Depth'' and ``NLU DMR Nodes'' are statistics based on the utterances for NLU.

We also compare DMR-FastFood with related open-source datasets in Table~\ref{tab:compare_data}. Though DMR-FastFood is not the biggest dataset, it has more turns per dialogue, longer utterance content, and more explicit annotations of negation and conjunction.

\section{Comparison of DMR with AMR and Dialogue-AMR}
\label{appendix:comparison}
In this section, we make detailed comparisons of DMR with AMR and Dialogue-AMR. 

\subsection{DMR vs AMR}
The key differences between DMR and AMR are as follows:

\begin{itemize}
    \item AMR is a sentence-level meaning representation, and it is designed for general purpose. DMR is proposed for task-oriented dialogue (TOD), so there are special contents in DMR designed for TOD, such as Intent, Entity, and cross-turn coreference. These contents are not included in AMR.
    \item AMR has more than 8k pre-defined predicates and is intended to capture as much semantics as possible. While DMR is task-driven, it only captures the contents that are related to the dialogue task.
    \item AMR abstracts surface forms to concepts, while DMR keeps the surface forms (\texttt{<lexical\_value>}) for entities, because the abstraction would depend on applications. For example, ‘big mac’ in the fast food domain is a type of burger, DMR abstracts it to ‘Burger’ entity, and keeps the surface form which could be important for the downstream module, e.g. the entity linking module, to find the exact burger item.
    \item DMR ontology is featured with inheritance hierarchy which serves for reusability across domains.
\end{itemize}

\subsection{DMR vs Dialogue-AMR} 
Dialogue-AMR is an extension of AMR, so the differences described above also apply in the case of Dialogue-AMR. Further, we highlight two more differences between DMR and Dialogue-AMR as follows. 

First, they serve for different applications. Dialogue-AMR is designed for human-robot interaction, and focuses on the mapping of natural language instructions to specific robotic control commands (e.g movement), whereas DMR focuses on the human interaction with software systems and their underlying business logic. The intents in DMR map to API calls, and entities map to pre-defined (compositional) data structures.

Second, there are multiple structural differences between them. Dialogue-AMR is an extension of AMR, which captures the illocutionary force (speech act), tense and aspect, spatial information in addition to AMR. In contrast, DMR is motivated by compositional intent-slot structures, and borrows key ideas of AMR to make it generalizable across applications in a task-oriented dialogue system. The main structural differences are: 
\begin{itemize}
    \item Dialogue-AMR captures information about speech acts, aspects, spatial information for human-robot interaction, as well as other concepts (defined in AMR) for general purpose. In contrast, DMR captures task-related intents, entities, relations defined in the ontology of a TOD application and ignores irrelevant contents.
    \item DMR represents coreferences across turns, which are common in TOD, while Dialogue-AMR does not.

\end{itemize}






\section{Details about the Inheritance Hierarchy in DMR}
\label{appendix:inheritance}
Inheritance Hierarchy is designed for better domain extensibility by reusing and expanding existing ontology.  Extensibility and reusability of domain ontology are essential for chatbot development, given the fact that a task-oriented dialogue system generally supports large amounts of applications, and each application often serves multiple domains. 

The effectiveness and efficiency of  the inheritance hierarchy are rooted in the observation that despite  different domains, some intents and entities   can be abstracted into one intent or entity.  For instance,  the intent of ordering is a general intent not only for fast food ordering, but also for flight tickets,  taxis, and  meeting room ordering. Similarly, the ‘PaymentIntent’, and ‘ConfirmationIntent’, can also be reused across domains.  On the other hand, different domains imply different requirements. Items to order in fast-food domain (food, drink) are different from those in the flight domain (tickets), which requires necessary adaptation on ontology. 

To this end, the inheritance hierarchy allows  inheriting from existing intents and entities, which allows  preserving the arguments of their parent and  modifying new arguments as needed at the same time. It is noted that the constraints about the arguments can be  also derived through inheritance hierarchy. In this  way, DMR arguments would  help reduce much less effort in designing ontology for new domains while maintaining high extensibility.  For instance, given pre-defined 'OrderIntent' and 'Item', ordering intent in the flight domain can reuse the properties of ‘OrderIntent’ as 'OrderFligthtTicketIntent', e.g. they can order multiple items;  they can only order orderable things. And ‘FlightTicket’ can inherit from ‘Item’ like ‘FoodItem’, because they are all orderable and countable, etc.

\section{Experiment Settings}
\label{appendix:exp}







\subsection{Hyperparameters}
The hyperparameters for the reported results are as follows.

\paragraph{DMR Parsing}
The DMR parsing models share the following hyperparameters: Adam optimizer, batch size 10, greedy search, and 10 training epochs. Others are listed in Table~\ref{tab:parsing_hyper_params}.
\begin{table}[h]
    \centering
    \setlength{\tabcolsep}{.5mm}
    \begin{tabular}{lll}
        \toprule
        Model     & Hyper-parameter         & Value \\
        \hline
        \multirow{6}*{\makecell[c]{BiLSTM           \\+GloVe}} & Embedding size & 300 \\
                  & Encoder layers          & 2     \\
                  & Decoder layers          & 2     \\
                  & Hidden size             & 512   \\
                  & Dropout                 & 0.1   \\
                  & Learning rate           & 1e-3  \\
        \hline
        \multirow{4}*{\makecell[c]{RoBERTa          \\-base}}  & Decoder layers & 2 \\
                  & Decoder attention heads & 4     \\
                  & Decoder hidden size     & 768   \\
                  & Learning rate           & 3e-5  \\
        \hline
        BART-base & Learning rate           & 1e-5  \\
        \bottomrule
    \end{tabular}
    \caption{Hyperparameters for DMR parsing models.}
    \label{tab:parsing_hyper_params}
\end{table}

\paragraph{Coreference Resolution}
\label{coref_hyp}
The hyperparameters for the GNN-based coreference resolution model (GNNCoref) are listed in Table~\ref{tab:coref_hyper_params}. Note that the value of threshold $\beta$ is tuned on the development set during training. Specifically, during each validation step, pick values in list $[0.01, 0.02, \cdots, 0.09]$ as the thresholds and calculate the accuracies, choose the threshold with the highest accuracy as $\beta$. GNNCoref model has 2.5M parameters.

\begin{table}[h]
    \centering
    \begin{tabular}{ll}
        \toprule
        Hyper-parameter & Value     \\
        \hline
        R-GCN Layers    & 3         \\
        Dropout         & 0.2       \\
        Epoch           & 20        \\
        Batch size      & 10        \\
        Learning rate   & 1e-3      \\
        Hidden size     & 100       \\
        Activation      & LeakyReLU \\
        $\beta$         & 0.89      \\
        \bottomrule
    \end{tabular}
    \caption{Hyper parameters for GNNCoref.}
    \label{tab:coref_hyper_params}
\end{table}
\subsection{Training}
\label{hyper-params}

All the models are trained on one NVIDIA Tesla T4 16G GPU. The training time of the DMR parsing models is 2.3 hours, 2.4 hours and 33 minutes for BiLSTM+GloVe, RoBERTa-base and BART-base respectively. And the GNNCoref model can be trained within 9 minutes.

\begin{table*}[t]
    \centering
    \begin{tabular}{lccccc}
         \toprule
                                   & TOP    & SB-TOP & TreeDST          & SMCalFlow       & DMR-FastFood    \\
                \hline
                Dialogue           & -      & -      & 27,280           & \textbf{41,517} & 7,194           \\
                Utterance/Dialogue & -      & -      & 7.1              & 4.1             & \textbf{18.5}   \\
                Annotated Turns    & 44,783 & 64,815 & \textbf{167,507} & 155,923         & 70,328          \\
                Utterance Length   & -      & 8.1    & 7.6              & 10.1            & \textbf{10.2}   \\
                Reference          & -      & 3,154  & 9,609            & \textbf{45,520} & 7,846           \\
                Negation           & -      & -      & -                & -               & \textbf{557}    \\
                Conjunction        & -      & -      & -                & 9,885           & \textbf{16,087} \\
                \bottomrule
    \end{tabular}
    \caption{Comparison of DMR-FastFood with related datasets that have tree/graph-structured representation for task-oriented dialogue systems.}
    \label{tab:compare_data}
\end{table*}

\begin{figure*}
    \centering
    \begin{subfigure}[b]{0.3\textwidth}
        \includegraphics[width=\textwidth]{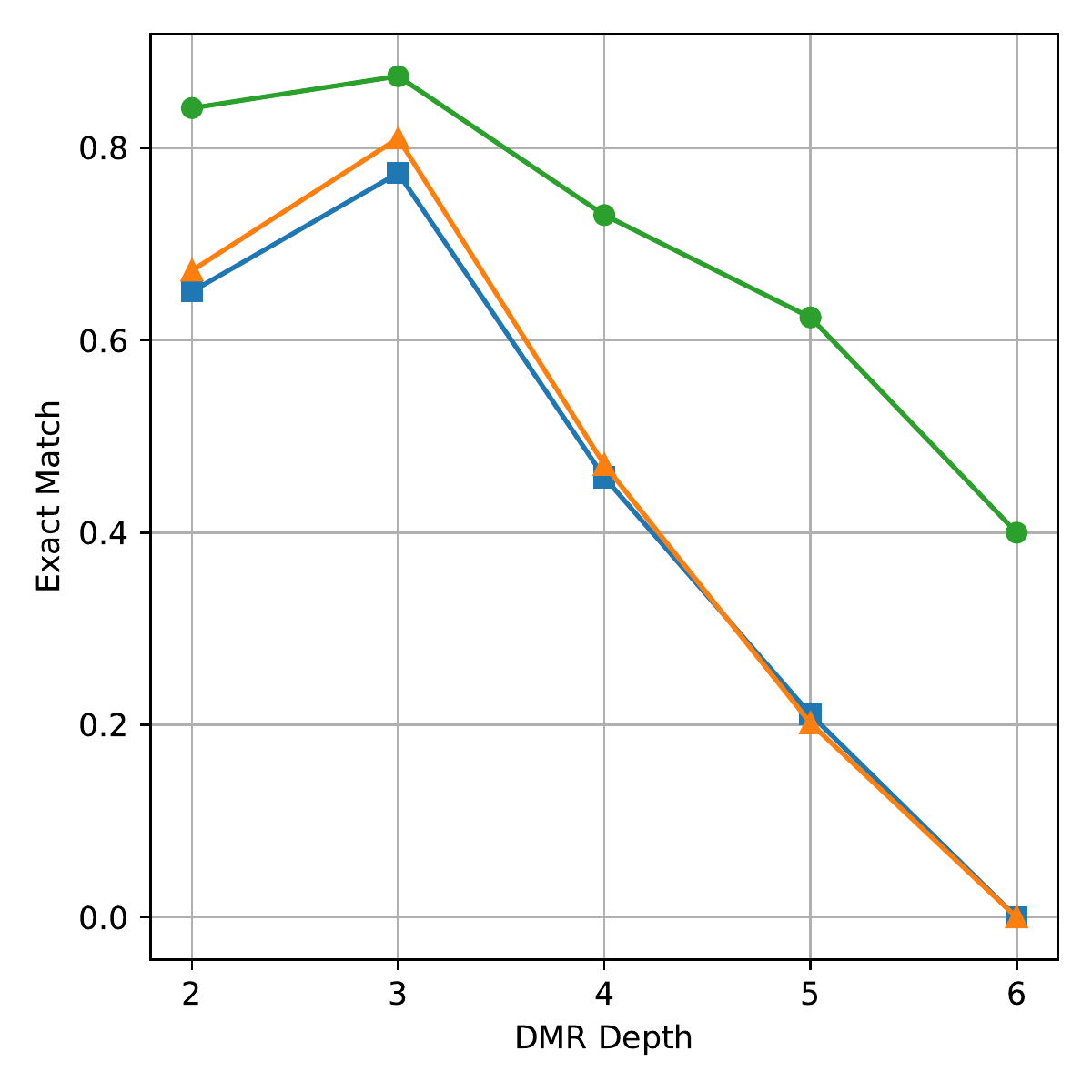}
        \caption{}
        \label{fig:dmr_depth_results}
    \end{subfigure}
    \begin{subfigure}[b]{0.3\textwidth}
        \includegraphics[width=\textwidth]{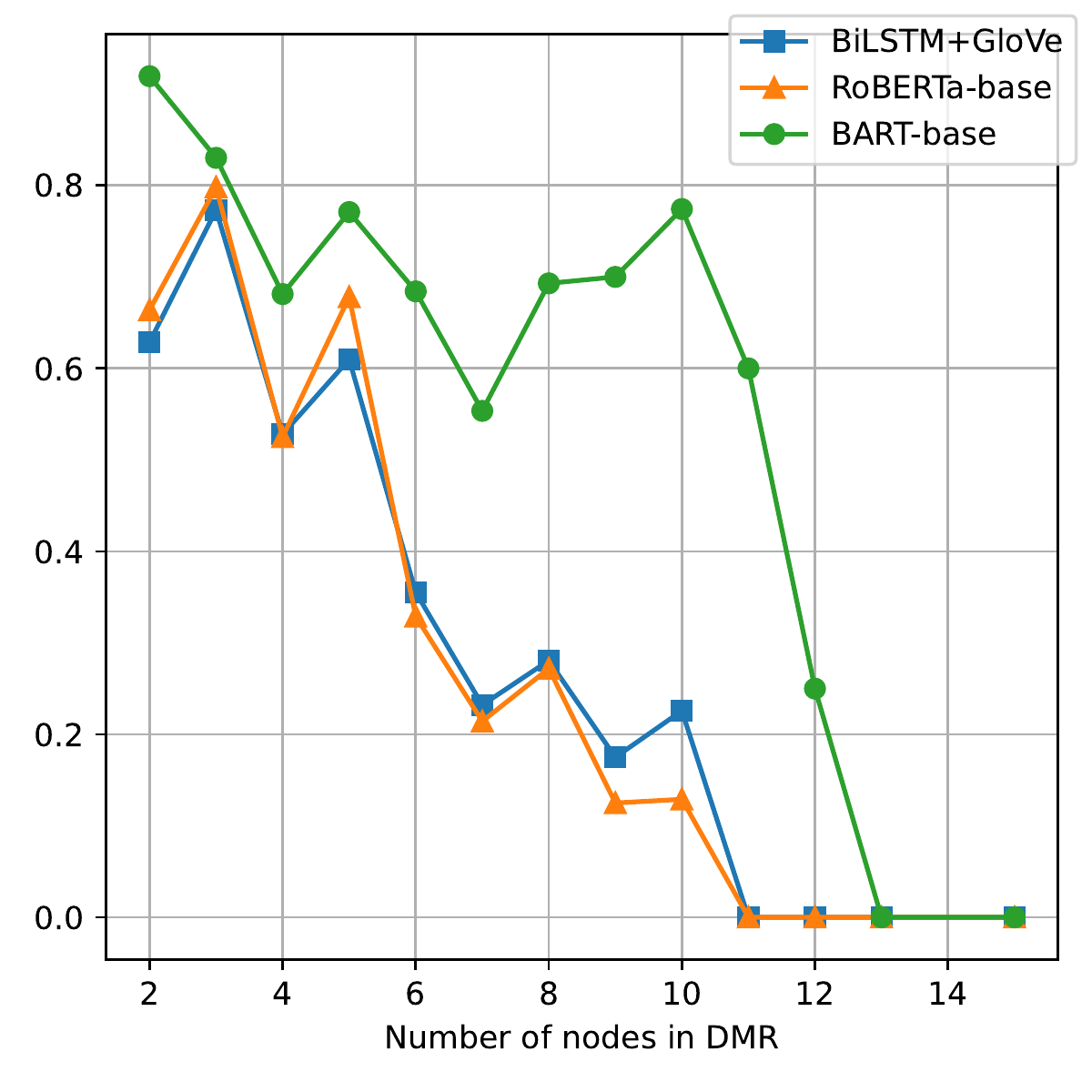}
        \caption{}
        \label{fig:dmr_num_nodes_results}
    \end{subfigure}
    \begin{subfigure}[b]{0.3\textwidth}
        \includegraphics[width=\textwidth]{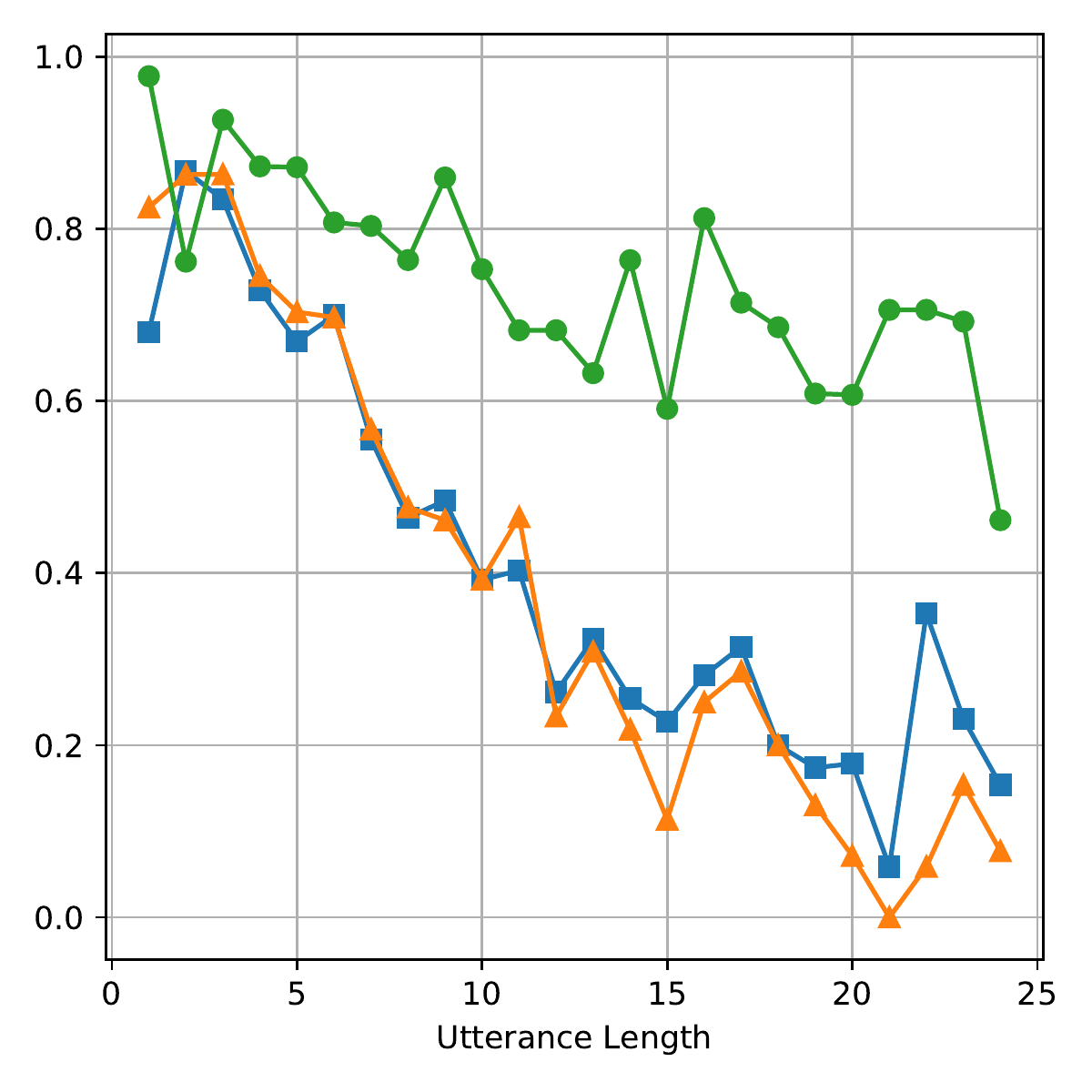}
        \caption{}
        \label{fig:utterance_length_results}
    \end{subfigure}
    \caption{DMR Parsing model performance analysis by a) DMR Depth b) Node Numbers, and c) Utterance Length.}
    \label{fig:my_label}
\end{figure*}

\section{More Analysis of the DMR Parsing model}
\label{appendix:analyze_parsing}

To investigate factors that affect the performances, we analyze the DMR Parsing model from four aspects: 1) the size of the dialogue context, 2) the depth of the target DMR, 3) the number of nodes in the target DMR, and 4) the content length of the utterance.

\paragraph{Context Size}

For each DMR Parsing model, we vary the context size $c$ from 0 to 3. The comparison of the results are listed in Table~\ref{tab:parsing_results_context_size}.\begin{table}[h]
    \centering
    \begin{tabular}{lccc}
        \toprule
        Model                       & $c$ & Dev set        & Test set       \\
        \hline
        \multirow{4}*{BiLSTM+GloVe} & 0   & 65.41         & 64.63          \\
                                    & 1   & 65.57          & 65.75          \\
                                    & 2   & \textbf{66.54} & \textbf{65.93} \\
                                    & 3   & 66.37         & 65.79          \\
        \hline
        \multirow{4}*{RoBERTa-base} & 0   & 65.93          & 66.29         \\
                                    & 1   & \textbf{69.24}         & \textbf{68.23}          \\
                                    & 2   & 67.31        & 67.49 \\
                                    & 3   & 63.20          & 62.55          \\
        \hline
        \multirow{4}*{BART-base}    & 0   & 80.89         & 81.99         \\
                                    & 1   & \textbf{82.56} & \textbf{83.39} \\
                                    & 2   & 82.18          & 83.26          \\
                                    & 3   & 82.14          & 82.79          \\
        \bottomrule
    \end{tabular}
    \caption{DMR Parsing results with different context size $c$.}
    \label{tab:parsing_results_context_size}
\end{table} The models get the best results with one or two context utterances, indicating the DMR Parsing is highly context-dependent.

The following analysis are based on the test set results reported in Table~\ref{tab:parsing_results}.

\paragraph{DMR Depth}
We compare the performance of different DMR parsing models at different DMR depths in Figure~\ref{fig:dmr_depth_results}. The performance drops for all models as the depth gets larger. Thus, the DMR depth is a good indicator of task complexity.

\paragraph{Node Number}
The more nodes in a DMR, the longer sequence to predict. Results in Figure~\ref{fig:dmr_num_nodes_results} in line with our intuition. Moreover, we see the BART-base model performs much better than the other two on large DMR targets, indicating that a well-pretrained decoder is critical for long sequence generation.

\paragraph{Utterance Length}
We plot the DMR parsing results for different utterance lengths in Figure~\ref{fig:utterance_length_results}. As expected, the models perform worse on longer utterances, and the BART-based model outperforms others substantially on these challenging test cases. This may be due to the correlation of the length of utterance and target sequence: in general, the more people say, the more information to be delivered. Thus, this result is consistent with Figure~\ref{fig:dmr_num_nodes_results}.

\end{document}